\documentclass[letterpaper, 10 pt, conference]{ieeeconf}  

\IEEEoverridecommandlockouts                              

\usepackage{graphics} 
\usepackage{epsfig} 
\usepackage{times} 
\usepackage{amsmath} 
\usepackage{amssymb}  

\usepackage{color}
\usepackage[ruled,vlined,linesnumbered]{algorithm2e}
\usepackage{algorithmic}
\usepackage{subcaption}

\newcommand{\modelname}{\text{HYPER} }
\newcommand{\modelnamenospace}{\text{HYPER}}

\usepackage{sidecap}
\newcommand{\rev}[1]{{#1}}

\title{\LARGE \bf
HYPER: Learned Hybrid Trajectory Prediction \\via Factored Inference and Adaptive Sampling
}

\author{Xin Huang$^{1}$, Guy Rosman$^{2}$, Igor Gilitschenski$^{2}$, Ashkan Jasour$^{1}$,\\ Stephen G. McGill$^{2}$, John J. Leonard$^{1,2}$, Brian C. Williams$^{1}$
\thanks{$^{1}$Computer Science and Artificial Intelligence Laboratory, Massachusetts Institute of Technology, Cambridge, MA 01239, USA
        {\tt\footnotesize xhuang@csail.mit.edu }}%
\thanks{$^{2}$Toyota Research Institute, Cambridge, MA 02139, USA }%
}

\begin{document}

\maketitle
\thispagestyle{empty}
\pagestyle{empty}

\begin{abstract}
    Modeling multi-modal high-level intent is important for ensuring diversity in trajectory prediction. 
    Existing approaches explore the discrete nature of human intent before predicting continuous trajectories, to improve accuracy and support explainability.
    However, these approaches often assume the intent to remain fixed over the prediction horizon, which is problematic in practice, especially over longer horizons.
    To overcome this limitation, we introduce \modelnamenospace, a general and expressive hybrid prediction framework that models evolving human intent. By modeling traffic agents as a hybrid discrete-continuous system, our approach is capable of predicting discrete intent changes over time. We learn the probabilistic hybrid model via a maximum likelihood estimation problem and leverage neural proposal distributions to sample adaptively from the exponentially growing discrete space. The overall approach affords a better trade-off between accuracy and coverage.
    We train and validate our model on the Argoverse dataset, and demonstrate its effectiveness through comprehensive ablation studies and comparisons with state-of-the-art models.
\end{abstract}

\section{Introduction}
	Predicting future trajectories of traffic agents is a key task for autonomous vehicles. This task is challenging due to multi-modal human intent. 
	There is an inherent trade-off between accurately representing the distribution of trajectories and covering the diversity of potential intents~\cite{makansi2019overcoming,Thiede2019,huang2020diversitygan}.
	
	Several recent works address the trade-off explicitly using a multi-stage approach~\cite{deo2018multi,guan2020generative,zhao2020tnt,zhang2020map,gilles2021home}. First, they infer high-level human intent, such as driving maneuvers and goal locations, to provide task-specific coverage, such as maximizing the space covered by the sampled goals~\cite{gilles2021home}. Next, trajectories are generated conditioned on the intent. The models are trained to maximize the data likelihood to support prediction accuracy. They demonstrate great success in terms of prediction accuracy and coverage, and provide explainability in predicted trajectories. 
	However, the existing approaches often use a simplified intent model that assumes the intent is \textit{fixed} over time, to keep the prediction space reasonable. In practice, a traffic agent can change its intent (i.e. follow the lane, perform a lane change, and turn), especially over long horizons. 
	
	When accounting for the evolving discrete intent, the number of discrete modes grows \emph{exponentially} in the prediction horizon~\cite{hofbaur2004hybrid}.
	This is studied in the context of factored inference, e.g. by merging and pruning mode hypotheses~\cite{blom1988interacting,andersson1985adaptive} or sampling from the prediction space~\cite{blackmore2008combined,koller2013general}.
	In the domain of trajectory prediction, the exponentially growing discrete space can be eased by expanding the discrete predictions at a few selected points~\cite{jayaraman2021multimodal} or accounting for the most probable intent~\cite{kothari2021interpretable}; however, they may not provide sufficient accuracy and coverage in a multi-modal problem.

\begin{figure*}
	\vspace{2mm}
	\begin{minipage}{0.3\textwidth}
	    \centering
        \includegraphics[scale=0.5]{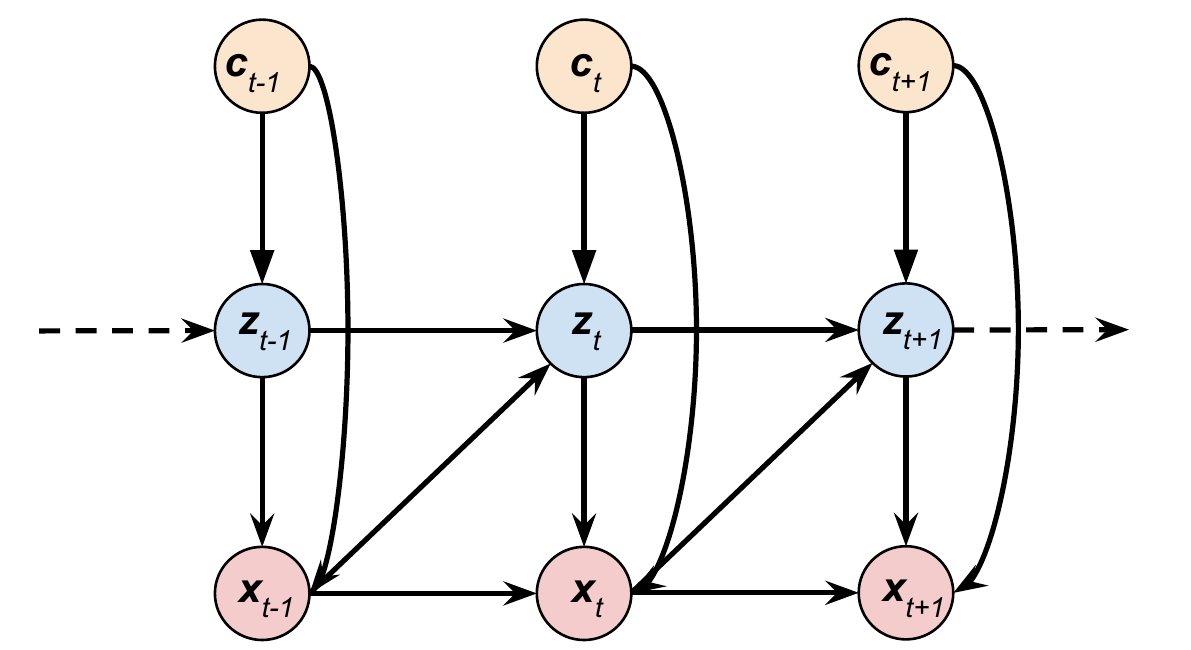}\\        (a)
    \end{minipage}
    \begin{minipage}{0.7\textwidth}
        \centering
        \includegraphics[scale=0.5]{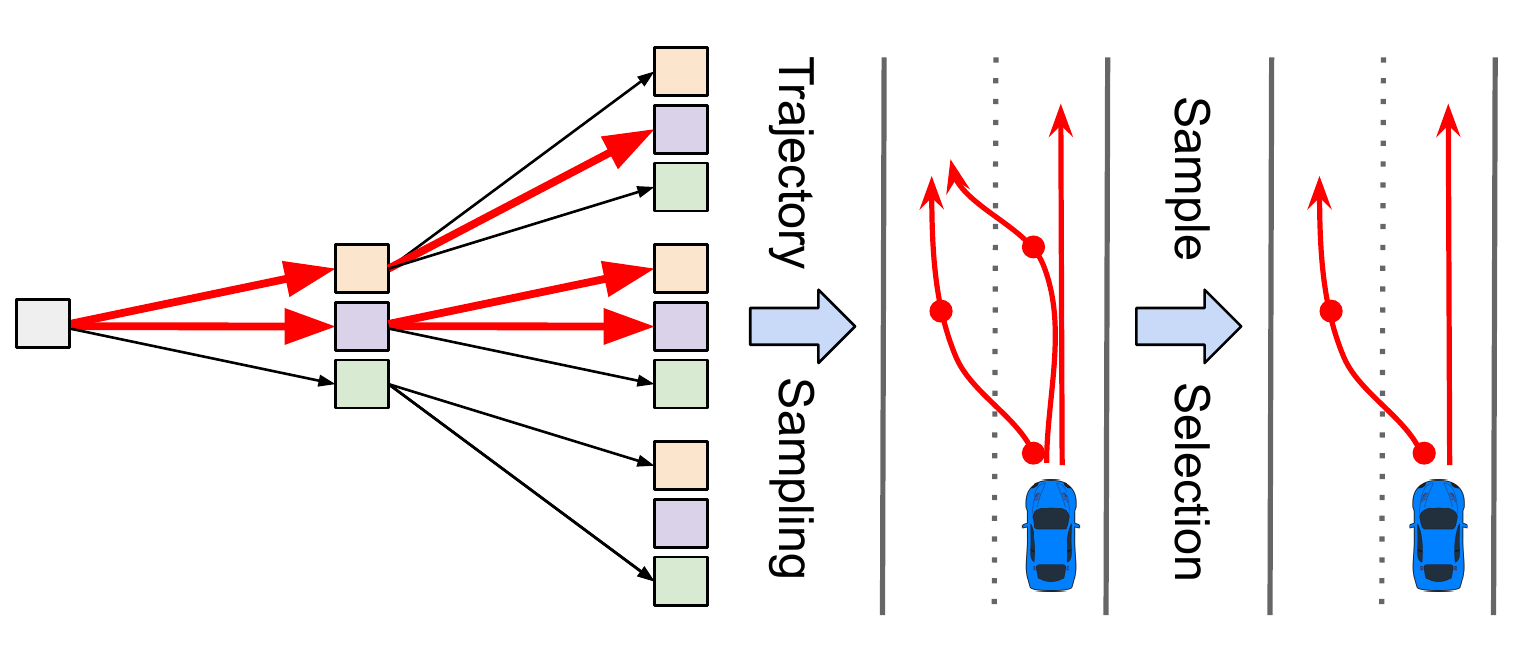}\\
        (b)
    \end{minipage}
    \caption{(a) Graphical model of a hybrid system representing traffic agents, where $\mathbf{z}$, $\mathbf{x}$, $\mathbf{c}$ represent discrete mode variables, continuous state variables, and context variables, respectively. Arrows indicate variable dependencies over time. (b) Overview of \modelnamenospace: Given a learned hybrid model, it leverages the learned proposal distribution to generate hybrid sequence samples (red arrows) from an exponentially growing space, which capture intent changes (red dots) over time and support coverage, and further chooses a small set of accurate and diverse trajectories using a sample selection algorithm.}
    \label{fig:overview}
    \end{figure*}

	We propose an approach that better captures both accuracy and coverage by explicitly modeling discrete intent sequences. Our approach, HYbrid trajectory PrEdictoR (\modelnamenospace), uses a learned probabilistic hybrid automata model, as illustrated in Fig.~\ref{fig:overview}(a), to jointly infer a sequence of high-level discrete modes when generating low-level trajectory predictions. We use neural proposal distributions \cite{gu2015neural} in the hybrid model and the farthest point sampling algorithm to obtain good coverage of trajectories with only a few samples, while preserving model accuracy.
	Our contributions are as follows: i) We formulate trajectory prediction as a general and expressive \emph{hybrid prediction problem} allowing an \emph{evolving discrete intent}, and learn a probabilistic hybrid automaton model as a deep neural network. 
	ii) \rev{We leverage a learned proposal function to sample \emph{adaptively} from an exponentially growing discrete space in the hybrid model to support both \emph{accuracy and coverage}, and utilize a sample selection technique to further improve prediction performance given limited samples.} 
	iii) \rev{We train and validate our model using a naturalistic driving dataset and perform detailed experiments to validate our hypothesis and demonstrate the effectiveness of our approach}.
	
\section{Related Work}
\label{sec:related_work}

\textbf{Multi-Modal Trajectory Prediction} Trajectory prediction has been studied extensively in the past few years. 
To account for uncertainty and multi-modality in prediction space, generative adversarial networks (GAN)~\cite{gupta2018social} and variational autoencoders (VAE)~\cite{lee2017desire,salzmann2020trajectron++} are used to generate multiple trajectory predictions by sampling a latent space. Several works have attempted to improve coverage of the possible outcomes~\cite{makansi2019overcoming,huang2020diversitygan,phan2020covernet,yuan2019diverse}, yet there is an inherent trade-off between accurately representing the trajectory distribution and covering a diverse set of intents~\cite{makansi2019overcoming,Thiede2019,huang2020diversitygan}.
To account for this trade-off between accuracy and coverage explicitly, hybrid models are proposed to classify discrete intent and generate continuous trajectories conditioned on the intent. The intent is defined by a variety of choices, including driving maneuvers~\cite{deo2018multi,richardos2020vehicle,hasan2021maneuver}, goal locations or waypoints~\cite{zhao2020tnt,gilles2021home,mangalam2020not,mangalam2020goals,liu2021multimodal,tran2021goal}, and target lanes~\cite{zhang2020map,kim2021lapred,song2021learning,narayanan2021divide}, etc. 
In these hybrid approaches, the intent is assumed to be fixed over time. 
In practice, however, the agent may change its intent over time, especially over a long horizon, or follow different intents to get to the same target location or lane. 
When accounting for evolving discrete intent, \cite{jayaraman2021multimodal} leverages a support vector machine to infer discrete intent over specific decision points, and \cite{kothari2021interpretable} proposes a discrete choice modeling approach to infer discrete anchors over time. Such models either expand discrete predictions at a few selected steps or predict the most-probable intent, to avoid dealing with the exponentially growing discrete space. In this work, we propose a general and expressive hybrid prediction framework that accounts for evolving intent by inferring a sequence of discrete modes over time, and predicting trajectories consistent with the mode sequence.

\textbf{Factored Inference} The discrete prediction space suffers from exponential growth as a function of the prediction horizon. This problem has been addressed in the context of factored inference, by approximating the intractable state space through pruning and sampling techniques. For instance,
multiple model estimation algorithms estimate the possible operational modes for a system, and filter states from an exponential number of hypotheses, by merging and pruning hypotheses~\cite{blom1988interacting,andersson1985adaptive}. 
Furthermore,~\cite{koller2013general} models the hybrid system through a hybrid Bayesian network, and proposes a sampling-based approximation algorithm to track hybrid states.
\rev{In parallel,~\cite{blackmore2007model,blackmore2008combined,timmons2020best,becker2019switching} model hybrid systems through a probabilistic hybrid automaton (PHA)~\cite{hofbaur2002mode} or a switching linear dynamical systems (SLDS)~\cite{linderman2017bayesian},} and apply efficient pruning, search, and sampling methods to maintain reasonable estimation performance. 
\rev{Existing factored inference methods often assume a linear system to obtain closed-form or trackable solutions, and do not fully utilize the relevant context information. In this work, we model a hybrid system as deep neural networks, which excel at modeling nonlinear agent dynamics and learning complex environment contexts such as map information.}

\textbf{Trajectory Sampling} 
\rev{Many trajectory prediction methods \cite{alahi2016social,gupta2018social,salzmann2020trajectron++} sample multiple predictions in parallel from a learned distribution. Our approach utilizes a sequential adaptive sampling technique to generate samples through a learned proposal distribution, conditioned on previously generated samples. This allows to provide more coverage with limited samples. The learned adaptive proposal function is inspired by a few ideas from sequential trajectory prediction sampling~\cite{park2018sequence}, sequential Monte Carlo~\cite{del2012adaptive,gu2015neural,naesseth2018variational}, and adaptive sampling~\cite{xu2011sequential,bardenet2014towards,xiao2018new}. Compared to existing sequential Monte Carlo methods, our proposal function is used to sample per time step, as opposed to sampling a full trajectory at once, to avoid a large proposal state.}

In many applications such as autonomous driving, only a small set of prediction samples can be afforded, as evaluating each sample for downstream tasks such as risk assessment is expensive~\cite{wang2020fast}. To select a limited number of candidates from all predicted samples,~\cite{huang2020diversitygan,yuan2019diverse} leverage diverse sampling techniques to choose semantically meaningful samples from a latent space;~\cite{zhao2020tnt} uses non-maximum suppression (NMS) to prune trajectories that are close to each other to improve coverage;~\cite{gilles2021home} proposes task-specific sub-sampling techniques towards optimizing the evaluation metrics. Similar to \cite{zhao2020tnt}, our approach selects samples directly over the predicted trajectories, offering better interpretability.

\section{Problem Formulation}
\label{sec:problem_formulation}
In this section, we introduce the hybrid system model used for \modelnamenospace, followed by a formal problem statement on learning this model.

\begin{figure*}
\vspace{2mm}
    \centering
    \includegraphics[scale=0.6]{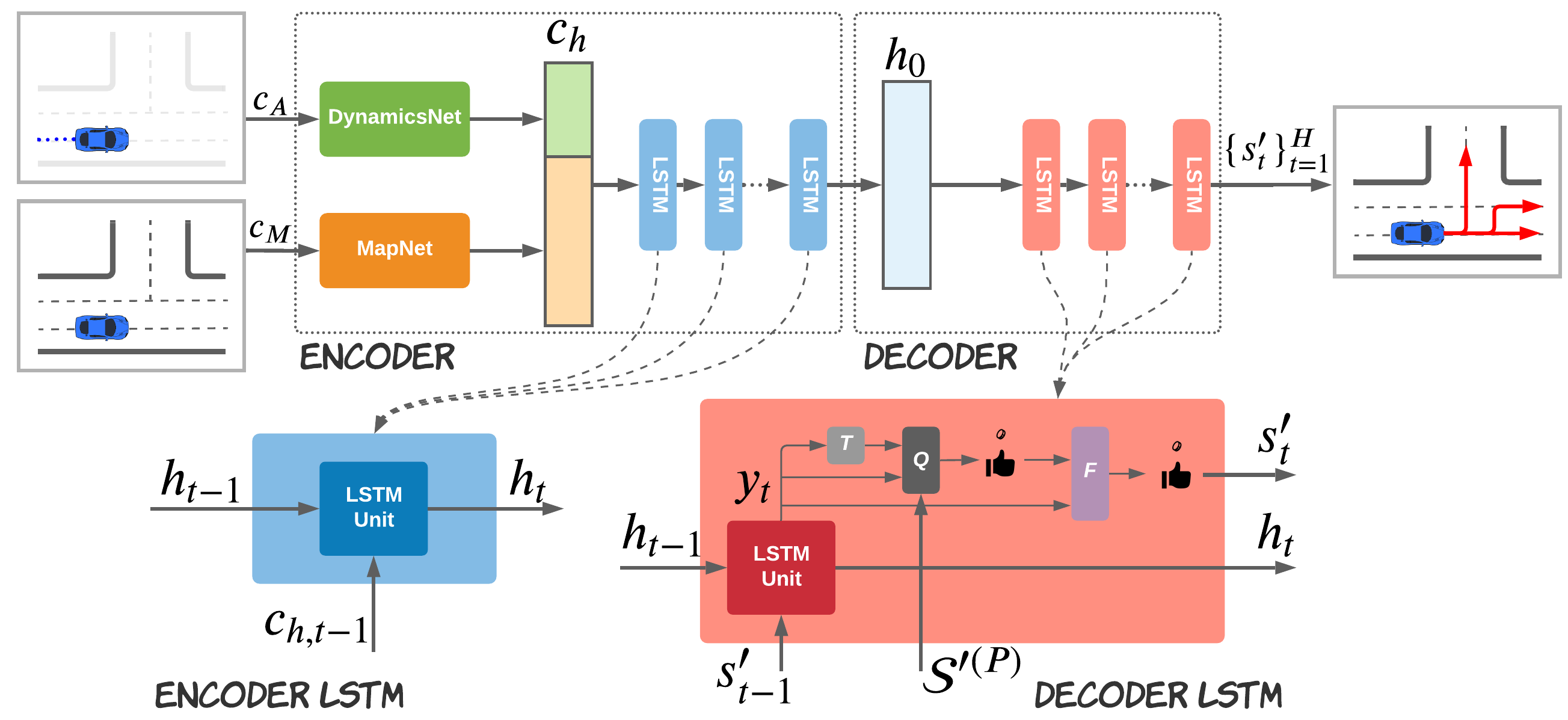}
    \caption{\rev{Overview of the proposed deep neural network.} The encoder encodes the context information, such as observed path $c_A$ and map $c_M$, and passes the combined encoded state $c_h$ through an LSTM network to obtain the hidden state vector $h_0$. The decoder is another LSTM that generates a sequence of hybrid states through a learned hybrid model, including a transition function $T$, a dynamics function $F$, and a proposal distribution $Q$ for improving coverage performance.}
    \label{fig:model_overview}
\end{figure*}

\subsection{Hybrid System Modeling}
\label{sec:pha}
We model a traffic agent as a \emph{probabilistic hybrid automaton} (PHA)~\cite{hofbaur2002mode}. \rev{Compared to hidden Markov model~\cite{krishnan2017structured}, the transitions in PHA have an autonomous property~\cite{blackmore2008combined}, i.e. the discrete mode evolution depends on the continuous state. This property provides better model capacity in mimicking the behavior of traffic agents.}
The PHA is a tuple $\mathcal{H}= \langle \mathbf{s}, \mathbf{w}, \mathit{F}, \mathit{T}, \mathbf{s}_0, \mathcal{Z}\rangle$\footnote{\rev{We use the lowercase bold symbols to denote both the set of variables and the vector, as in~\cite{blackmore2008combined}, and lowercase standard symbols to denote variable instantiations.}},
where $\mathbf{s} = \mathbf{x} \cup \mathbf{z}$ denotes the hybrid state variables -- $\mathbf{z}$ denotes the discrete mode with a finite domain $\mathcal{Z}$, and $\mathbf{x} \in \mathbb{R}^{n_x}$ denotes continuous state variables; $\mathbf{w}$ specifies the input/output variables, which consists of context variables $\mathbf{c}$, continuous observation variables $\mathbf{o}_x$, and discrete observation variables $\mathbf{o}_z$; \rev{$\mathit{F}: \mathcal{Z} \to \mathcal{F}$ specifies the continuous evolution of the automaton for
each discrete mode, in terms of a set of discrete-time difference equations $\mathcal{F}$ over the variables $\mathbf{x}$ and $\mathbf{c}$}; \rev{$\mathit{T}: \mathcal{Z} \to \mathcal{T}$ specifies the discrete evolution of the automaton for
each discrete mode, as a finite set of transition probabilities drawn from $\mathcal{T}$}; $\mathbf{s}_0$ denotes the initial hybrid state.

The dependencies of hybrid state variables in a PHA are depicted as a graphical model in Fig.~\ref{fig:overview}(a), where we omit the observation variables for simplicity. The state evolution is governed by the transition function $T$ and the dynamics function $F$, indicated by the three arrows going to $\mathbf{z}_t$ and to $\mathbf{x}_t$, respectively. 

\rev{Similar to existing trajectory prediction approaches, we assume that the discrete labels are observable at training time, defined as driving maneuvers~\cite{deo2018multi,richardos2020vehicle,hasan2021maneuver} or goal locations~\cite{zhao2020tnt,gilles2021home}. These labels can be obtained through auto-labelling or unsupervised clustering over continuous trajectories. While it is challenging to perfectly label driver intent, we show in the experiments that our model is robust to imperfect labels, and defer learning with hidden intent for future work.}

\subsection{Hybrid Model Learning}
\label{sec:hybrid_learning}
Given a set of observed discrete-continuous future agent states $O = (O_x, O_z)$\footnote{\rev{While the continuous state can be directly observed from perception systems, the discrete observation can be estimated from continuous observations, as discussed in Sec.~\ref{sec:pha}.}} and context states $C$, we want to learn a hybrid model parameterized by $\theta$ that maximizes the following data log likelihood~\cite{blackmore2007model}, as a maximum likelihood estimation (MLE) problem:
{\footnotesize
\begin{equation}
\begin{split}
    &\mathcal{L}_{\text{MLE}}(O, C) \\=& \sum_{o, c \in (O, C)} \log p(o| c; \theta)\\=& \sum_{o, c} \sum_{t=1}^H \log p_F(o_{x}^t | o_{x}^{t-1}, o_{z}^t, c; \theta) + \log p_T(o_{z}^t | o_{x}^{t-1}, o_{z}^{t-1}, c; \theta),    
\end{split}
\label{eq:ll}
\end{equation}}
where $o = \{(o_x, o_z)^t\}_ {t=1}^{t=H}$ is an observed future hybrid trajectory sequence with horizon $H$, $p_F$ is a Gaussian distribution over continuous states, $p_T$ is a categorical distribution over discrete modes. The prior for the observations is omitted as we assume that the observation noise is negligible, which is common in most trajectory prediction literature.


\section{Approach}
\label{sec:model}
In this section, we introduce our approach to learn a PHA-based encoder-decoder deep neural network model, as depicted in Fig.~\ref{fig:model_overview}. The encoder embeds the context information $c$ into a hidden state vector $h_0$, and the decoder samples a sequence of hybrid states $\{s'_t\}_{t=1}^H$ up to a finite horizon through a hybrid model, conditioned on $h_0$.
Given the learned model, we use the decoder to sample multiple predictions and apply farthest point sampling in the continuous trajectory space to generate a small set of predictions with good coverage, as visualized in Fig.~\ref{fig:overview}(b).

\subsection{Encoder}
In the encoder, we first encode the context information $c$, including the observed path of the target agent $c_A$ and the observed map data $c_M$, per time step in the past. The observed path at each step is encoded through a multi-layer perceptron (MLP). The map is encoded through a model based on~\cite{gao2020vectornet}, which takes the map input as a set of lane centerlines, and performs self-attention to pool the encoded states from all centerlines. Next, we run the encoded context state $c_h$ through an LSTM network, \rev{which is commonly used in handling sequential data in trajectory prediction~\cite{park2018sequence,deo2018multi,huang2020diversitygan}}, to get the hidden context state vector $h_0$ at the most recent observed step $t=0$.

\subsection{Decoder}
In the decoder, we generate a sequence of hybrid states, including discrete modes and continuous positions, using another LSTM network. At each time step, the LSTM unit takes as inputs the hidden state $h_{t-1}$ and the hybrid state sample $s'_{t-1}$ from the previous step, and outputs a new hidden state $h_t$ and an output state $y_t$. The output LSTM state $y_t$ is passed through a transition function $T$\footnote{\rev{By definition, the transition function takes the previous state and the context information directly as inputs, as in Fig.~\ref{fig:overview}(a). We follow existing sequential prediction models \cite{sutskever2014sequence,alahi2016social,gupta2018social} to read its inputs through an LSTM, and abuse $T$ to represent the auxiliary transition function that takes the LSTM output $y_t$.}}, modeled as an MLP layer that produces a categorical distribution $P_T(z_t)$ over discrete modes at time $t$ as logits. 
\rev{
Existing factored inference algorithms~\cite{koller2013general,blackmore2008combined} sample from $T$ to obtain discrete samples; however, naively sampling from $T$ may take a large number of samples to sufficiently cover the prediction space.
Therefore, we propose learning an additional \emph{proposal function} $Q$ on the top of the transition function, to sample the discrete mode for the \textit{task} of achieving better prediction accuracy and coverage. 
The proposal function takes input from i) the distribution from $T$, ii) the output LSTM state $y_t$, and iii) the set of previously generated trajectory sequence samples $\mathcal{S}'^{(P)}$, and outputs a categorical distribution over the next discrete modes as logits. The last input allows us to sample adaptively conditioned on previously generated samples. Each previous full trajectory sample is encoded through an MLP layer. The max pooling of all sample encodings is passed through an MLP layer before being fed into the proposal function. Compared to existing sequential Monte Carlo methods~\cite{del2012adaptive,gu2015neural,naesseth2018variational}, our proposal function is used to sample per time step, as opposed to sampling a full trajectory at once, to avoid a large proposal state.}

\rev{Given the output of $Q$, we leverage a Gumbel-softmax sampler to sample the next mode $z'_t$, whose real probability is obtained from $T$. The mode is concatenated with $y_t$ and fed into a dynamics function $F$ as an MLP layer that outputs the distribution of the continuous state $P_F(x_t)$. The distribution is parameterized as a Gaussian distribution with mean $\mu_{x'_t}$ and unit variance, which is chosen arbitrary for stable training and is assumed in models such as~\cite{zhao2020tnt}. We then sample a continuous state $x'_t$ from the distribution. The predicted hybrid sample $s'_t = (z'_t, x'_t)$ at time step $t$ and the hidden state $h_t$ are used to generate the sample in the next step, until the prediction horizon is reached.}

\rev{
\textbf{Transition Function vs. Proposal Function:}
Although both $T$ and $Q$ output a categorical distribution over modes, they serve different purposes. The transition function $T$ is part of the hybrid model defined in Sec.~\ref{sec:pha}, and it is used to compute the \textit{real} probability of a sample. For instance, given an observation of a future hybrid state sequence, we can compute its likelihood by plugging it into the LSTM model (i.e. through $T$ and $F$). The likelihood is the summation of the discrete log-likelihood from $P_T$ and continuous log-likelihood from $P_F$, as in Eq.~\eqref{eq:ll}. It allows us to faithfully optimize the hybrid model, by maximizing the log-likelihood given the ground truth future observations, as defined in Sec.~\ref{sec:hybrid_learning}.}

\rev{
On the other hand, the proposal function $Q$ determines which samples to generate, in order to improve prediction coverage in an exponentially growing space. It does not represent the true sample weight, which is determined by the transition function $T$. The weighted sample set allows us to cover the prediction distribution efficiently with only a few samples, which is advantageous to existing sampling-based methods that require a large number of samples to approximate the probability distribution. In Sec.~\ref{sec:result}, we show a use case of sample weights to quantify prediction accuracy as negative log-likelihood.
}

\subsection{Learned Proposal Distributions}
To train the proposal function for accuracy and coverage, we generate $K$ trajectory sequences $\{s'^{(k)} = (z'^{(k)}, x'^{(k)})\}_{k=1}^K$ sequentially from the decoder, and compute the min-of-$K$ L2 loss compared to the ground truth continuous observations $o_x$:
\begin{equation}
    \mathcal{L}_Q = \min_{k \in K} || x'^{(k)} - o_x ||_{2}^2.
\label{eq:coverage}
\end{equation}

There exist a few other options to learn the proposal function for coverage, such as maximizing entropy~\cite{zhao2019maximum}. In this paper, we focus on the task of improving the diversity of the continuous trajectories when guaranteeing prediction accuracy, and choose the min-of-$K$ L2 loss (or variety loss~\cite{gupta2018social}) that is widely used in the multi-modal trajectory prediction literature.
While it is possible to train the model with only the min-of-$K$ L2 loss to favor towards prediction coverage, as in \cite{gupta2018social,huang2020diversitygan}, it leads to a diluted probability density function compared to the ground truth \cite{Thiede2019}. Therefore, we choose to improve prediction coverage while ensuring accuracy, by introducing the data likelihood loss in Eq.~\eqref{eq:ll}. As a result, we can leverage the proposal distribution to generate representative samples, while obtaining the real probability of these samples from the transition function. To encourage the proposal distribution to be close to the transition distribution, we add a regularization loss on the L2 differences between the two distribution logits $\mathcal{L}_{\text{reg}}$:
\begin{equation}
    \mathcal{L}_{\text{reg}} = ||T_{\text{logits}} - Q_{\text{logits}}||_2^2.
\end{equation}

\subsection{Trajectory Sample Selection}
In many autonomous vehicle applications, we can only afford a small set of prediction samples, due to the non-trivial computational complexity of evaluating these samples for risk assessment~\cite{wang2020fast}. To further improve coverage and boost prediction performance with a limited budget on samples, we propose to use the farthest point sampling (FPS) algorithm~\cite{gonzalez1985clustering}. The algorithm selects trajectories that are far away from each other from samples generated from the proposal distribution, while maintaining their probabilities through the learned hybrid model. The algorithm works by selecting the next sample farthest away from the previously selected samples, in terms of the distance between end locations, with the first sample selected with the highest likelihood. FPS is able to capture the majority of distinct options thanks to its 2-optimal coverage property~\cite{gonzalez1985clustering}, as we show in Sec.~\ref{sec:fps} on how it captures diverse samples with the proposal distribution.

\subsection{Model Training and Inference}
In training time, we jointly train the hybrid model and the proposal distribution with the loss
\begin{equation}
    \mathcal{L} = -\mathcal{L}_{\text{MLE}} + \alpha \mathcal{L}_{Q} + \beta \mathcal{L}_{\text{reg}},
\label{eq:loss_all}
\end{equation}
where the MLE term (c.f. Eq.~\eqref{eq:ll}) is negated as a loss to minimize, and $\alpha$ and $\beta$ are the loss coefficients. 


At inference time, we i) sequentially call the hybrid model $M$ times with the proposal function to generate $M$ hybrid trajectory sequences, ii) compute their likelihoods based on the probabilities from the transition function and the dynamics function, and iii) perform FPS to select the final $N$ trajectory samples, and normalize the probabilities of each sample so that they sum up to 1.

\section{Experimental Results}
\label{sec:result}
In this section, we introduce the dataset and the model details, followed by a series of experiments demonstrating the effectiveness of our approach compared to baselines.

\subsection{Dataset and Model Details}
We train and validate \modelname on Argoverse v1.1~\cite{chang2019argoverse}, a widely used benchmark for single agent trajectory prediction. The data contains 324,557 segments of agent trajectories, including two seconds of observed trajectories and three seconds of trajectories to predict, sampled at $10Hz$, as well as map information such as lane centerlines. \rev{Despite having a short prediction horizon, almost 40\% of Argoverse data exhibits evolving intents with more than one label in three seconds, as we demonstrate in the experiments.}
We augment the dataset offline with discrete mode labels over time, defined as \texttt{stop, fast forward, slow forward, left turn, right turn}, depending on the velocity and angular changes differentiated from the trajectories. 

In the encoder, DynamicsNet is an MLP with 32 neurons; MapNet utilizes a similar structure as VectorNet~\cite{gao2020vectornet}; the encoder LSTM has a hidden size of 32 and an output dimension of 32. In the decoder, the transition function and the proposal function use a two-layer MLP with (32, 5) neurons followed by a softmax layer; the dynamics function is a two-layer MLP with (32, 2) neurons; the sampler is a Gumbel-Softmax sampler~\cite{jang2016categorical} that produces differentiable samples; the decoder LSTM has the same structure as the encoder LSTM. All MLPs are followed by ReLU and dropout layers with a rate of 0.1. 

The loss coefficients $\alpha$ and $\beta$ in Eq.~\eqref{eq:loss_all} are selected to be 1. The sizes of samples $K, M, N$ are selected to be 6, 50, and 6, respectively.
The model is optimized using Adam~\cite{kingma2014adam} and trained on a single NVIDIA Tesla V100 GPU, with a batch size of 16 and a learning rate of 1e-3.

The prediction performance is evaluated by minimum average displacement error (ADE) and final displacement error (FDE)~\cite{gupta2018social} in meters, where ADE measures the average distance between the predicted trajectory and the ground truth, and FDE measures the distance at 3 seconds. The minimum error comes from the best predicted sample. All statistics are collected in the Argoverse validation dataset. For sampling-based methods (i.e. Gumbel-softmax), we run them five times and take the average. We annotate the method used in our model with \textit{italics} font in the tables. 

\subsection{Ablation Study}
We perform two ablation studies to validate the effectiveness of the adaptive proposal distribution and sample selection through FPS. 

\subsubsection{Learned Adaptive Proposal Distribution}
We demonstrate the contribution of the learned proposal distribution by comparing different options of discrete intent proposal functions, including i) the learned transition function $T$ \rev{(i.e. setting $\alpha$ and $\beta$ to 0)}, ii) the non-adaptive proposal function that with no access to the other samples, and iii) our proposed proposal function $Q$ that samples adaptively by considering previously generated samples. The results are summarized in Table.~\ref{table:discrete}, in which we obtain 6 samples \emph{without} further sample selection, and compute the errors of the best sample. We observe that the proposal functions, learned to optimize the minimum errors, result in better metrics compared to the discrete function, especially through adaptive sampling. \rev{In addition, the regularization term $\mathcal{L}_{reg}$ with a coefficient $\beta = 1.0$ helps stabilize training and avoid overfitting.}

\begin{table}
\vspace{2mm}
\centering
\begin{tabular}{lcccc}
          & \multicolumn{2}{c}{1 Second} & \multicolumn{2}{c}{3 Seconds} \\
Discrete Function & minADE           & minFDE          & minADE           & minFDE           \\
\hline \hline 
Transition         & 0.45             & 0.62            & 1.19             & 2.43             \\
Proposal (non-Adapt.)         & 0.44             & 0.48            & 1.00             & 1.92             \\
\textit{Proposal (Adaptive)}          & \textbf{0.33}             & \textbf{0.44}            & \textbf{0.86}             & \textbf{1.68}             \\
\hline \\   
\end{tabular}
\caption{Min-of-6 errors using different discrete function choices. Our proposed adaptive proposal function achieves the lowest errors.}
\label{table:discrete}
\end{table}

\begin{table*}
\vspace{2mm}
\centering
\begin{tabular}{lcccccc}
          & \multicolumn{2}{c}{6 / 6 samples} & \multicolumn{2}{c}{6 / 30 samples} & \multicolumn{2}{c}{6 / 50 samples} \\
Selection Method & minADE           & minFDE          & minADE           & minFDE & minADE           & minFDE           \\
\hline \hline 
Proposal + Random          & 0.86             & 1.68            & 1.02             & 2.27     & 0.98            & 2.12        \\
Proposal + Most-likely          & 0.86             & 1.68          & 1.10             & 2.45     & 1.13            & 2.51        \\
Proposal + NMS (2m)         & 0.86             & 1.68          & 0.76             & 1.38     & 0.73            & 1.30        \\
Proposal + NMS (4m)         & 0.86             & 1.68         & 0.80             & 1.57  & 0.78            & 1.49 \\
\textit{Proposal + FPS}         & \textbf{0.86}             & \textbf{1.68}        & \textbf{0.74}             & \textbf{1.30}     & \textbf{0.72}            & \textbf{1.26}        \\
Transition + FPS         & 1.19             & 2.43            & 1.06             & 2.00  & 1.03            & 1.96 \\ \hline\\
\end{tabular}
\caption{Min-of-6 ADE/FDE over 3 seconds using different sample selection methods. FPS achieves the best performance by selecting 6 samples generated from the proposal distributions.}
\label{table:subsampling}
\end{table*}

\subsubsection{Trajectory Sample Selection}
\label{sec:fps}
We validate the effectiveness of our sample selection method, FPS, by comparing it with a few standard options, including i) a random sampler picking samples based on their weights, ii) a most-likely sampler that selects the top likely samples, similar to selecting the most-probable intent in~\cite{kothari2021interpretable} and best-$k$ enumeration in~\cite{timmons2020best}, iii) a sampler based on non-maximum suppression (NMS), as used in~\cite{zhao2020tnt}, which selects samples greedily by finding the next sample that is distant enough from existing samples given a threshold. For a fair comparison, the distance measure in NMS is the same as FPS based on final locations, and we empirically choose 2 different distance thresholds (2 meters and 4 meters) to select the next sample. If the number of valid NMS samples is smaller than $N$, we select the remaining samples randomly. We also compare to the option of applying FPS over the samples generated from the transition function, to verify that the proposal function generates better samples.

In the study, we first generate $M \in \{6, 30, 50\}$ samples using the proposal distribution (or the discrete distribution for the bottom row), and select $N=6$ samples. The results are summarized in Table~\ref{table:subsampling}. When $M=N=6$, no subsampling occurs. 
When $M > N$, a random sampler and a most-likely sampler do not improve the errors, as selecting only the most likely samples leads to worse errors, since trajectory prediction is a multi-modal problem. As $M$ grows, the most-likely sampler acts similar to a maximum likelihood estimator, and exhibits inferior results as the problem is multi-modal.
NMS improves results but is limited by a fixed distance threshold: when the threshold is small (i.e. 2 meters), it fails to provide enough coverage in cases where the predicted samples are very far away; when the threshold is large (i.e. 4 meters), the number of valid samples can be smaller than $N$.
On the other hand, FPS reduces the errors the most, by finding the 6 samples that provide both accuracy and coverage. When $M$ is larger than 50, the error reduction is small for both NMS and FPS. We note that without the learned proposal distribution, FPS does not achieve the same results, as the discrete structure is not explored efficiently by the samples generated from the transition function. 

\subsection{Quantitative Results}
We compare our full model with a number of representative baselines, including i) \textbf{DESIRE}~\cite{lee2017desire} that utilizes a conditional VAE model to generate trajectory prediction samples from a latent space; \rev{ii) \textbf{DiversityGAN}~\cite{huang2020diversitygan} that predicts diverse samples by learning a latent space in a GAN model such that the prediction samples with different semantic meanings are far away in that latent space. This baseline is similar to other diverse sampling works that improve coverage through a learned latent space~\cite{yuan2019diverse,weng2021ptp};} iii) \textbf{MultiPath}~\cite{chai2019multipath} that learns the trajectory modalities as a set of anchors and predicts trajectories through anchor classification and offset regression; iv) \textbf{TNT}~\cite{zhao2020tnt} that first infers discrete target locations and second predicts target-conditioned trajectories to support multi-modality.
\rev{In addition, we introduce a few variants of our models to validate our hypothesis, including v) \textbf{SingleMode} that assumes a single mode and only samples from the continuous distribution;  vi) \textbf{\modelnamenospace-Linear} that uses only linear layers in the decoder to simulate a linear dynamic system as in existing factored inference literature that assumes linear dynamics; vii) \textbf{\modelnamenospace-Coverage} that is trained with only the task-specific \emph{coverage} loss, defined in Eq.~\eqref{eq:coverage} -- it shares the same spirit as~\cite{kothari2021interpretable} that optimizes for the minimum loss; viii) \textbf{\modelnamenospace-Perturb} that is trained on a dataset in which we randomly perturb 5\% of discrete labels.}

We use the metrics reported in~\cite{zhao2020tnt}, and present the comparison in Table~\ref{table:stoa}. \rev{\modelname outperforms all baselines that assume a fixed intent over time (i-iv), ignore discrete structure in the model (v), or presume linear dynamics (vi). In order to demonstrate its robustness with noisy discrete labels, we randomly perturbed 5\% of discrete labels and observe that our model trained on the perturbed data achieves similar results (viii).}
We further improve the minADE metric with a variant, \modelnamenospace-Coverage, that is solely trained towards optimizing this metric, but sacrifices accuracy measured by the negative log-likelihood metric (NLL)~\cite{ivanovic2019trajectron}. Our method, on the other hand, allows for the trade-off between accuracy and coverage.

\begin{table}
\centering
\begin{tabular}{lccc}
Model & NLL & minADE           & minFDE           \\
\hline \hline 
DESIRE~\cite{lee2017desire}        & - & 0.92             & 1.77             \\
DiversityGAN~\cite{huang2020diversitygan}        & - & 1.13             & 2.20             \\
MultiPath~\cite{chai2019multipath}  & -      & 0.80             & 1.68             \\
TNT~\cite{zhao2020tnt}  & -     & 0.73             & 1.29             \\ \hline 
\textit{SingleMode}    & 78.46      & 0.87            & 2.00             \\ 
\textit{\modelnamenospace-Linear}    & 45.86      & 0.79            & 1.38             \\ 
\textit{\modelnamenospace-Coverage}    & 34.54      & \textbf{0.66}            & 1.27             \\ 
\textit{\modelnamenospace-Perturb}    & 31.02      & 0.71            & 1.27             \\ 
\textit{\modelname}     & \textbf{30.87}     & 0.72             & \textbf{1.26}             \\
\hline \\   
\end{tabular}
\caption{NLL and min-of-6 ADE/FDE over 3 seconds compared to baseline models. Our model balances between accuracy and coverage, with a variant (\modelnamenospace-Coverage) trained on the coverage task achieving the lowest minADE.}
\label{table:stoa}
\end{table}

\begin{table*}
\vspace{2mm}
\centering
\begin{tabular}{lcccccc}
          & \multicolumn{3}{c}{1 Second} & \multicolumn{3}{c}{3 Seconds} \\
Model & minADE           & minFDE     & minDER       & minADE           & minFDE   & minDER        \\
\hline \hline 
ManeuverLSTM~\cite{deo2018multi}         & 0.41             & 0.52     & 5.70\%       & 1.06             & 1.94      & 11.01\%       \\
\textit{\modelname}          & \textbf{0.32}             & \textbf{0.40}   & \textbf{5.18\%}         & \textbf{0.80}             & \textbf{1.47}     & \textbf{7.65\%}        \\
\hline \\   
\end{tabular}
\caption{Compared to ManeuverLSTM, \modelname achieves better results in both discrete and continuous error metrics.
}
\label{table:maneuverlstm}
\end{table*}

\begin{figure*}
	\begin{minipage}{0.24\textwidth}
	    \centering
        \includegraphics[scale=0.16]{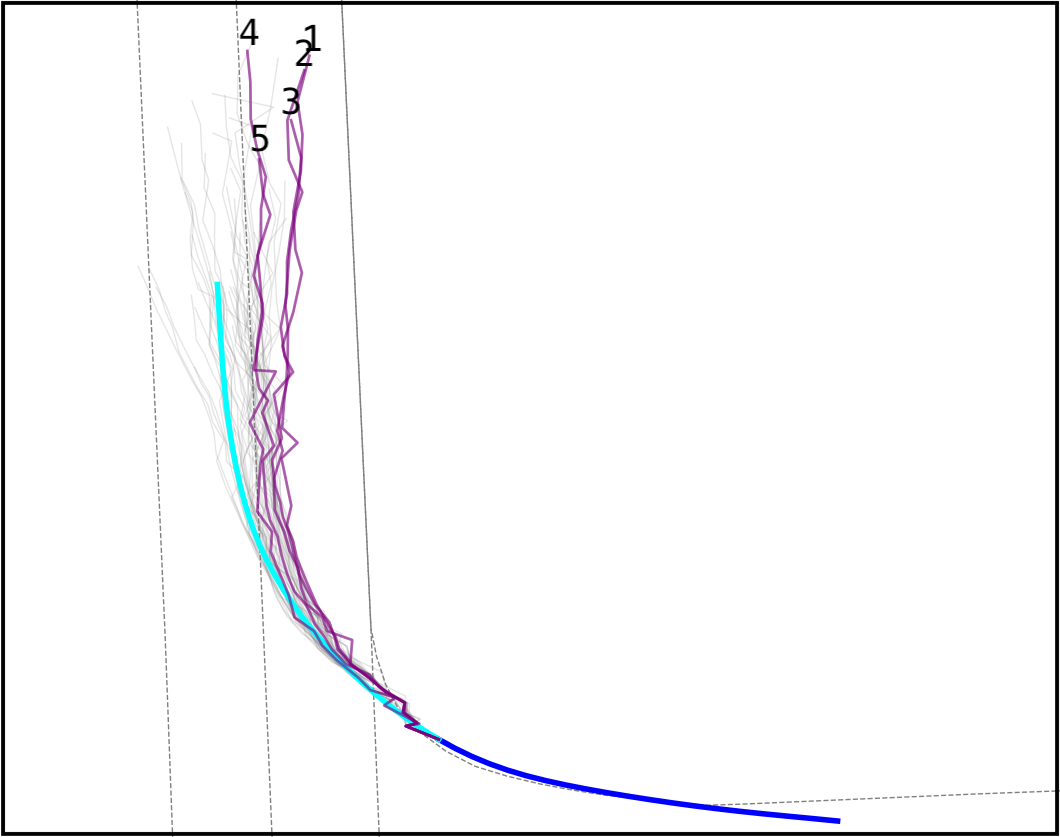}\\
        (a)
    \end{minipage}
    \begin{minipage}{0.24\textwidth}
	    \centering
        \includegraphics[scale=0.16]{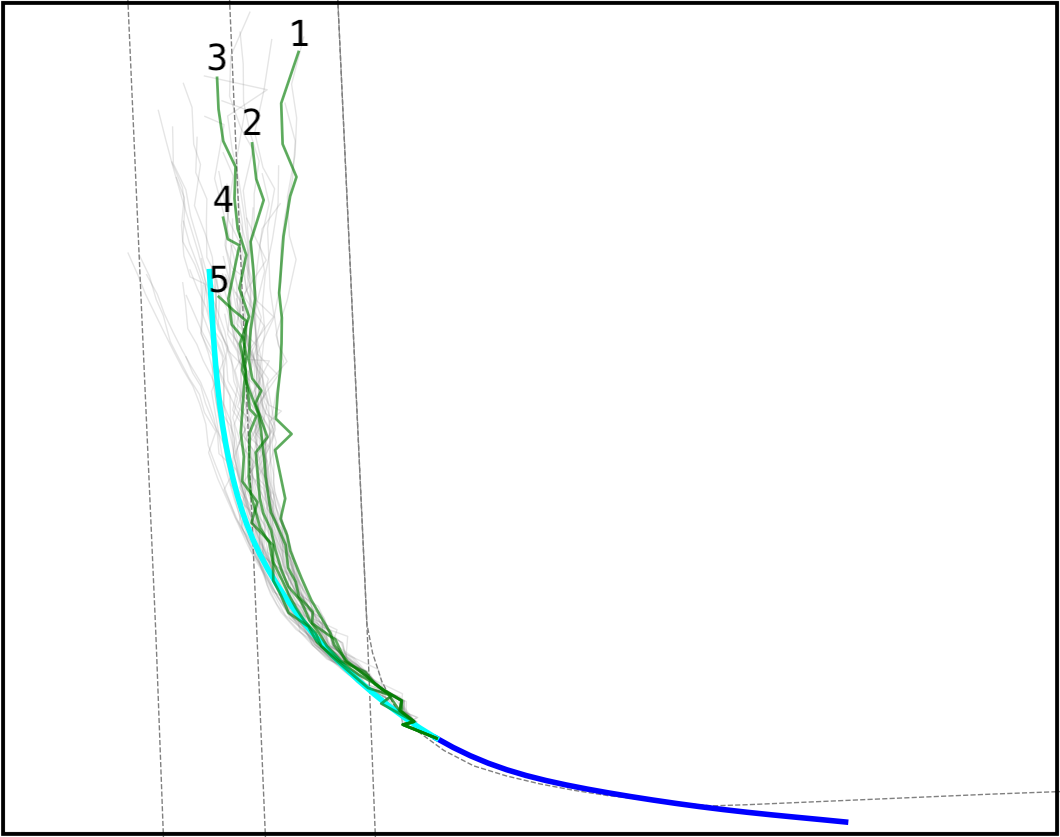}\\
        (b)
    \end{minipage}
    \begin{minipage}{0.24\textwidth}
	    \centering
        \includegraphics[scale=0.16]{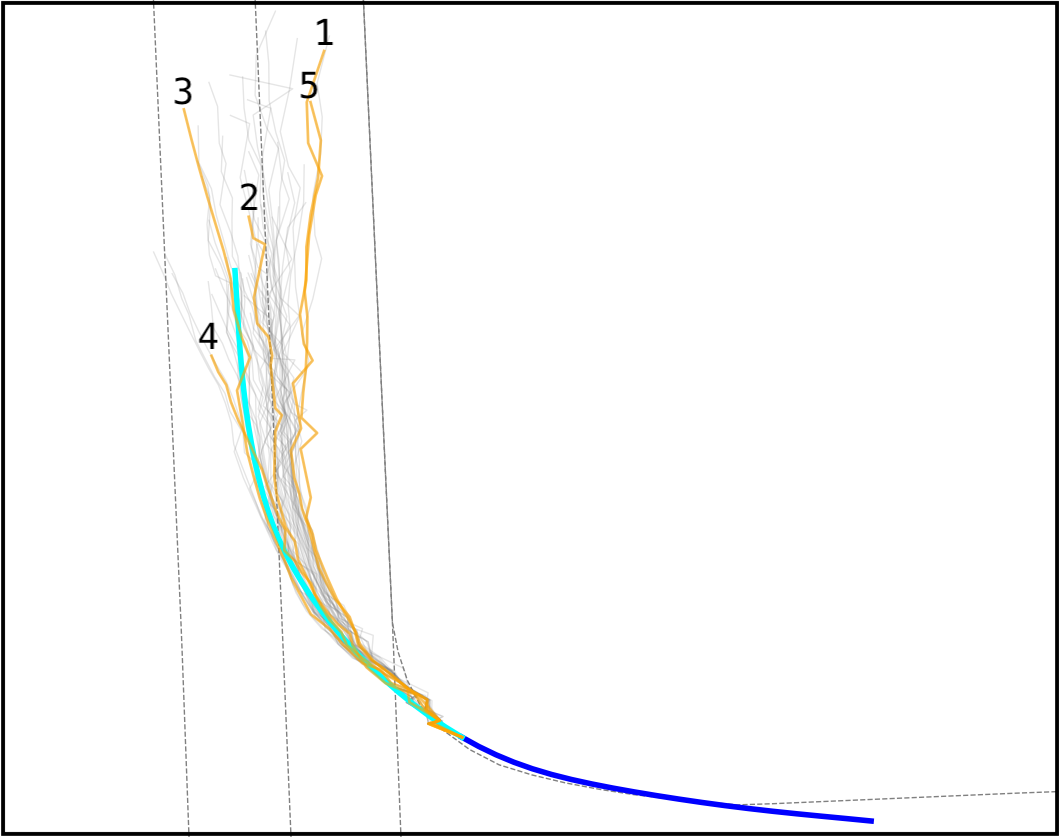}\\
        (c)
    \end{minipage}
    \begin{minipage}{0.24\textwidth}
	    \centering
        \includegraphics[scale=0.16]{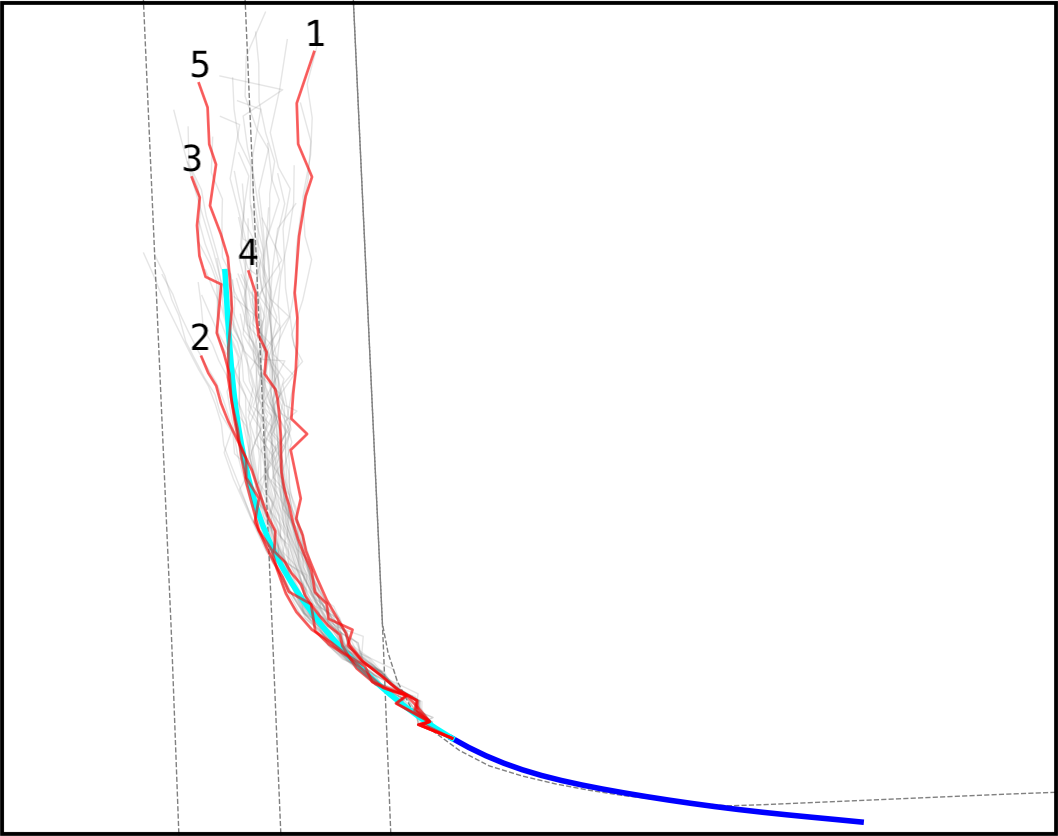}\\
        (d)
    \end{minipage}
    \caption{Sample selection using different methods: (a) most-likely (purple), (b) NMS (2m) (green), (c) NMS (4m) (orange), (d) FPS (red). Ground truth past and future trajectory are in blue and cyan. Predicted samples in grey. Numbers indicate the sampling order. FPS achieves the best coverage.}
    \label{fig:sampling}
    \end{figure*}
    
\begin{figure*}
	\begin{minipage}{0.49\textwidth}
	    \centering
        \includegraphics[scale=0.33]{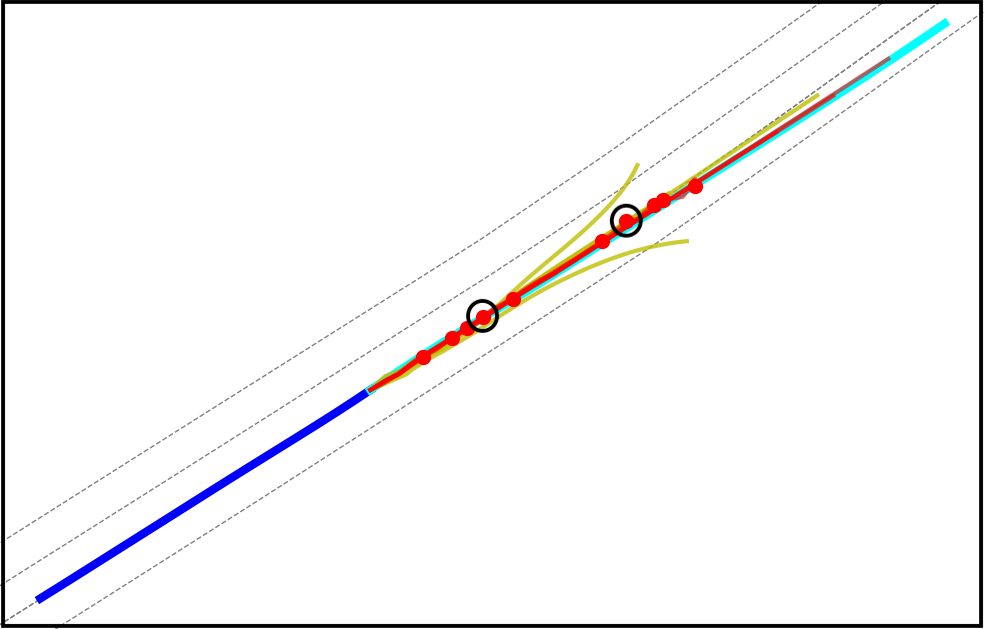}\\
        (a)
    \end{minipage}
    \hfill
    \begin{minipage}{0.49\textwidth}
	    \centering
        \includegraphics[scale=0.33]{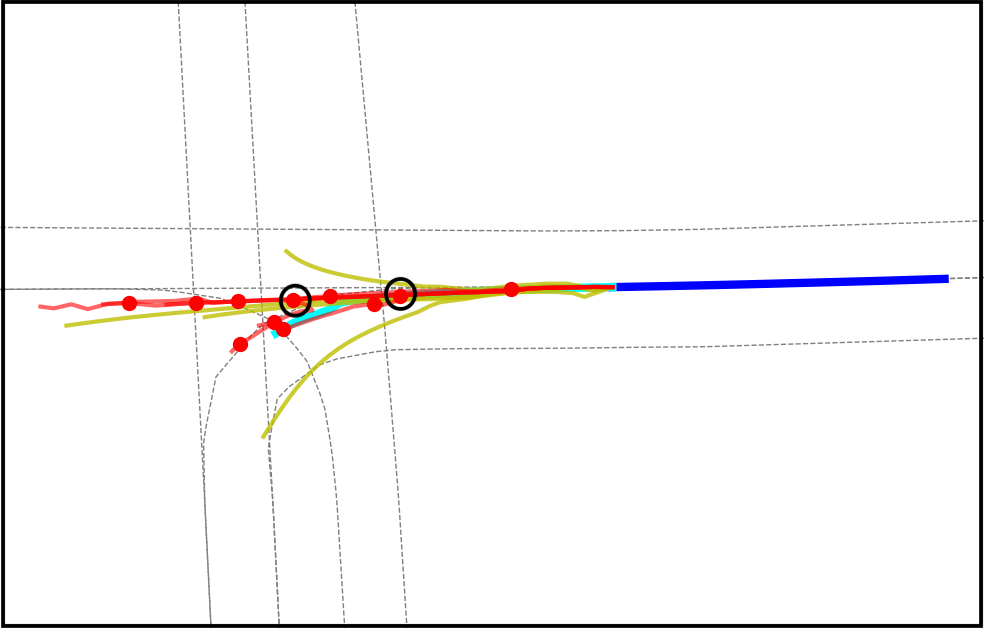}\\
        (b)
    \end{minipage}

\caption{Comparison between \modelname (red) and ManeuverLSTM (olive) predictions, where ground truth future trajectories are in cyan, and red dots depict where the mode changes in \modelname predictions. (a) In a lane change scenario, \modelname identifies multiple time slots, highlighted in black circles, to change the intent. (b) In a turning scenario, \modelname predicts a fast turn and a slow turn, highlighted in black circles, that is transitioned from a forward maneuver. In both scenarios, ManeuverLSTM has worse accuracy due to the assumption of fixed maneuvers over time.}
    \label{fig:comparison}
    \end{figure*}

The work that is closest to ours in spirit is \textbf{ManeuverLSTM}~\cite{deo2018multi}. It models driving modes explicitly as maneuvers labeled from trajectory data, and assumes the maneuver is fixed over time. For a fair comparison, \rev{we use the same model, training process, and definition of maneuvers as used in our model, except forcing each sample to have a fixed mode over the prediction horizons for ManeuverLSTM}. We use five samples for comparison given the number of maneuvers defined for ManeuverLSTM. In addition to the standard metrics, we introduce min-of-$K$ discrete error rate (minDER) that measures the percentage of wrong predictions in discrete states for the best predicted sample, to quantify the discrete prediction accuracy and coverage. Table~\ref{table:maneuverlstm} demonstrates that our model outperforms this baseline by a large margin, in both continuous and discrete metrics, by supporting evolving maneuver intent and utilizing a proposal function to explore the intent space.

\subsection{Qualitative Results}
In Fig.~\ref{fig:sampling}, we present a qualitative example to demonstrate the effectiveness of FPS. Fig.~\ref{fig:sampling}(a) shows the most likely examples selected based on the predicted likelihood, which favor the option to follow the middle lane or merge to the right lane. On the other hand, NMS selects more diverse samples, but suffers from a fixed distant threshold -- when the threshold is small (i.e. 2 meters), it does not have enough coverage; when the threshold is large (i.e. 4 meters), it does not have enough valid samples to select and has to resort to random samples. In Fig.~\ref{fig:sampling}(c), NMS (4m) finds only four valid samples and chooses the fifth one randomly that is close to the first one, failing to provide better coverage. As a more robust and threshold-free alternative, FPS finds diverse options more effectively.

In Fig.~\ref{fig:comparison}, we show two examples that demonstrate the advantage of supporting evolving driving modes. Fig.~\ref{fig:comparison}(a) depicts a lane change scenario, where the ground truth future trajectory follows the lane for a few seconds and then performs a lane change. Our model (predictions in red) infers the mode change successfully and predicts a few options on when to change, as highlighted by the red dots circled in black. Fig.~\ref{fig:comparison}(b) depicts a turning scenario in which the ground truth future trajectory follows the lane first and then performs a left turn. Again, our model identifies the mode change and predicts an early turn and a late turn (see red dots in black circles), which improve prediction accuracy. On the other hand, ManeuverLSTM (predictions in olive) only predicts a single option for each maneuver, ignoring the fact that modes may change in the future sequence. This leads to worse accuracy and coverage. For instance, in Fig.~\ref{fig:comparison}(a), ManeuverLSTM predicts sharp lane changes that are far away from ground truth future trajectory, due to its assumption on a fixed intent.


\section{Conclusion}
\label{sec:conclusion}
In conclusion, we present a general and expressive hybrid prediction model that accounts for evolving discrete modes in the future trajectory. The model leverages learned proposal functions and the farthest point sampling algorithm to select a small number of accurate and diverse samples from an exponential space. The effectiveness of our model is validated in the Argoverse dataset, through both quantitative and qualitative experiments.


\bibliographystyle{IEEEtran}
\bibliography{ref} 

\iftrue
\newpage
\appendix
\section{Additional Experiment Details}
\textbf{Discrete Mode Labeling} 
We augment the Argoverse dataset with discrete mode labels, defined as \texttt{stop, fast forward, slow forward, left turn, right turn}, using the following auto-labeling procedure: First, we compute the step-wise velocity magnitude and angular changes from trajectories. Second, at each time step, if the heading change is greater than a threshold $\theta$, the mode is \texttt{left turn}; else if the heading change is smaller than $-\theta$, the mode is \texttt{right turn}; else if the velocity magnitude is greater than $v_F$,  the mode is \texttt{fast forward}; else if the velocity magnitude is greater than $v_S$,  the mode is \texttt{slow forward}; else the mode is \texttt{stop}. We empirically set the thresholds $\theta, v_F, v_S$ to be 2.0, 1.0, and 0.05, respectively.

We run a Gaussian process filter to smooth the noisy trajectories prior to auto-labeling the maneuvers, using \textit{sklearn.gaussian\_process.GaussianProcessRegressor} library, with a \textit{Matern} kernel and an alpha value of 0.1 and default setting for the remaining parameters.

\section{Additional Model Details} 
\textbf{MapNet} 
We implement MapNet in our encoder based on VectorNet~\cite{gao2020vectornet}, with a few modifications. First, the MLPs in MapNet have a hidden dimension of 32, to be consistent with the remaining MLPs used in our model. Second, we add heading tangent values as additional inputs to the node feature. Third, we improve the auxiliary node completion task from reconstructing hidden node features to reconstructing explicit node inputs such as positions and headings. While the second and third modifications slightly improve the performance compared to a vanilla VectorNet, the performance improvement is mainly attributed to our learned proposal functions and sample selection scheme.

\section{Additional Qualitative Examples}
In the following, we introduce examples with a larger view to demonstrate the advantage of \modelnamenospace, in predicting evolving intents to improve accuracy and coverage, compared to ManeuverLSTM that assumes a fixed intent.

\newpage
\subsection{Lane follow and change}
\modelname (top) predicts multiple time slots to perform a lane change after following the lane for a few seconds and achieves better accuracy than ManeuverLSTM (bottom), which predicts a single option for the right turn by assuming fixed intent.

\begin{figure}[b!]
   \centering
        \includegraphics[trim=100 50 100 100,clip,scale=0.3]{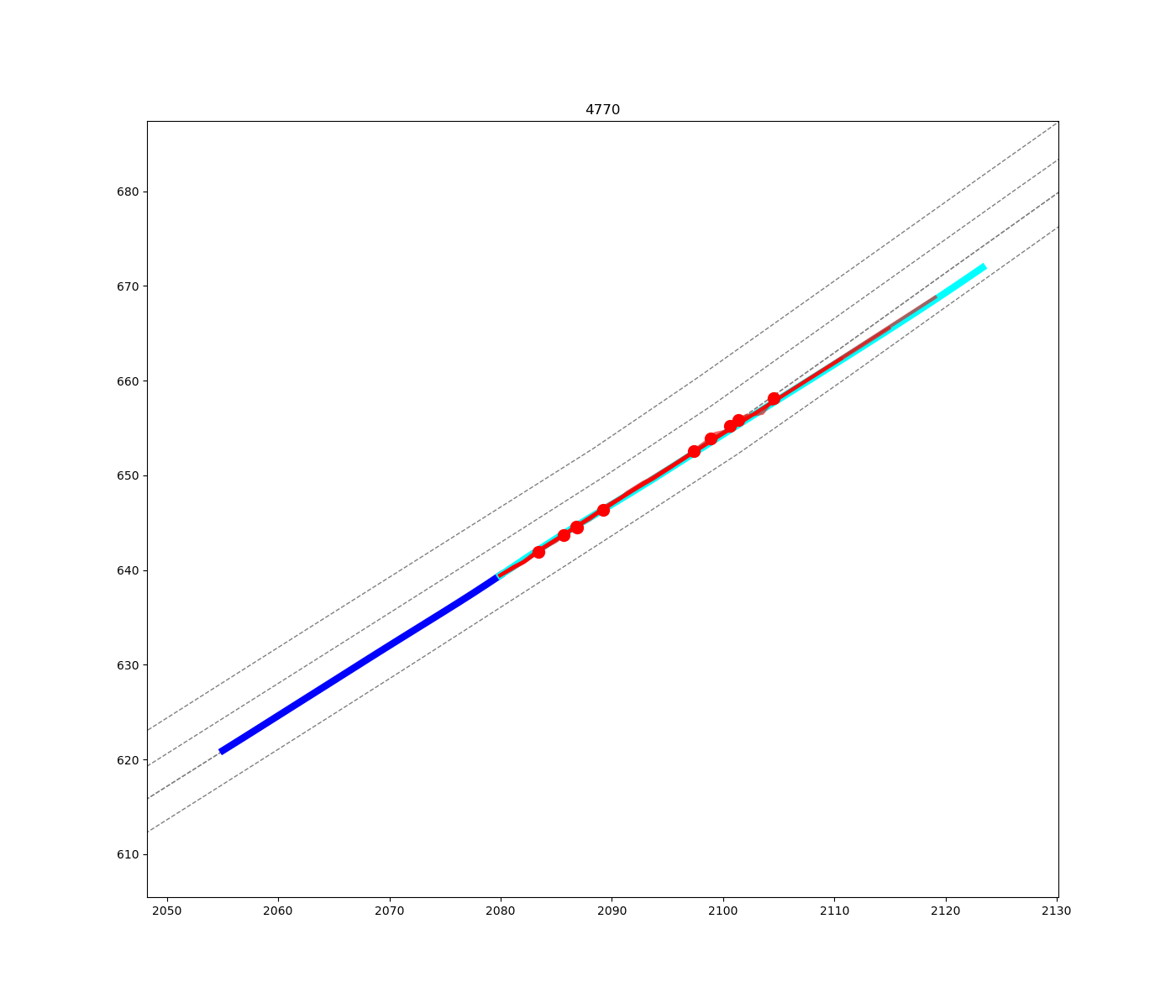}
        \includegraphics[trim=100 50 100 100,clip,scale=0.3]{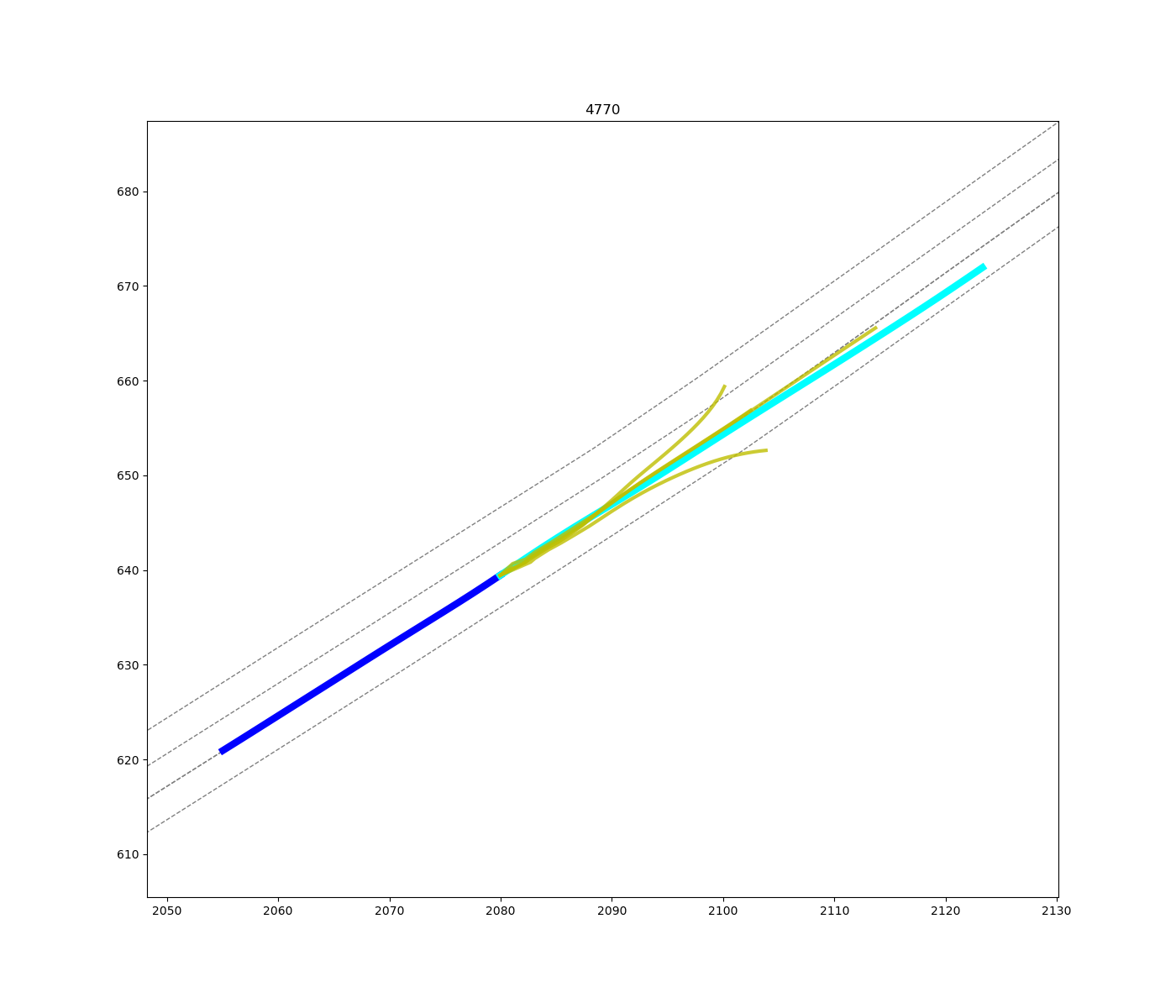}
\caption{Lane follow and change. Ground truth past and future trajectory are in blue and cyan. Top: Predictions from \modelname in red. Red dots depict the mode changes in predictions. Bottom: Predictions from ManeuverLSTM in olive.}
    \end{figure}

\newpage
\subsection{Lane change and follow}
\modelname (top) predicts the agent to change to the left lane and then follow that lane and improves accuracy.

\begin{figure}[b!]
   \centering
        \includegraphics[trim=100 50 100 100,clip,scale=0.3]{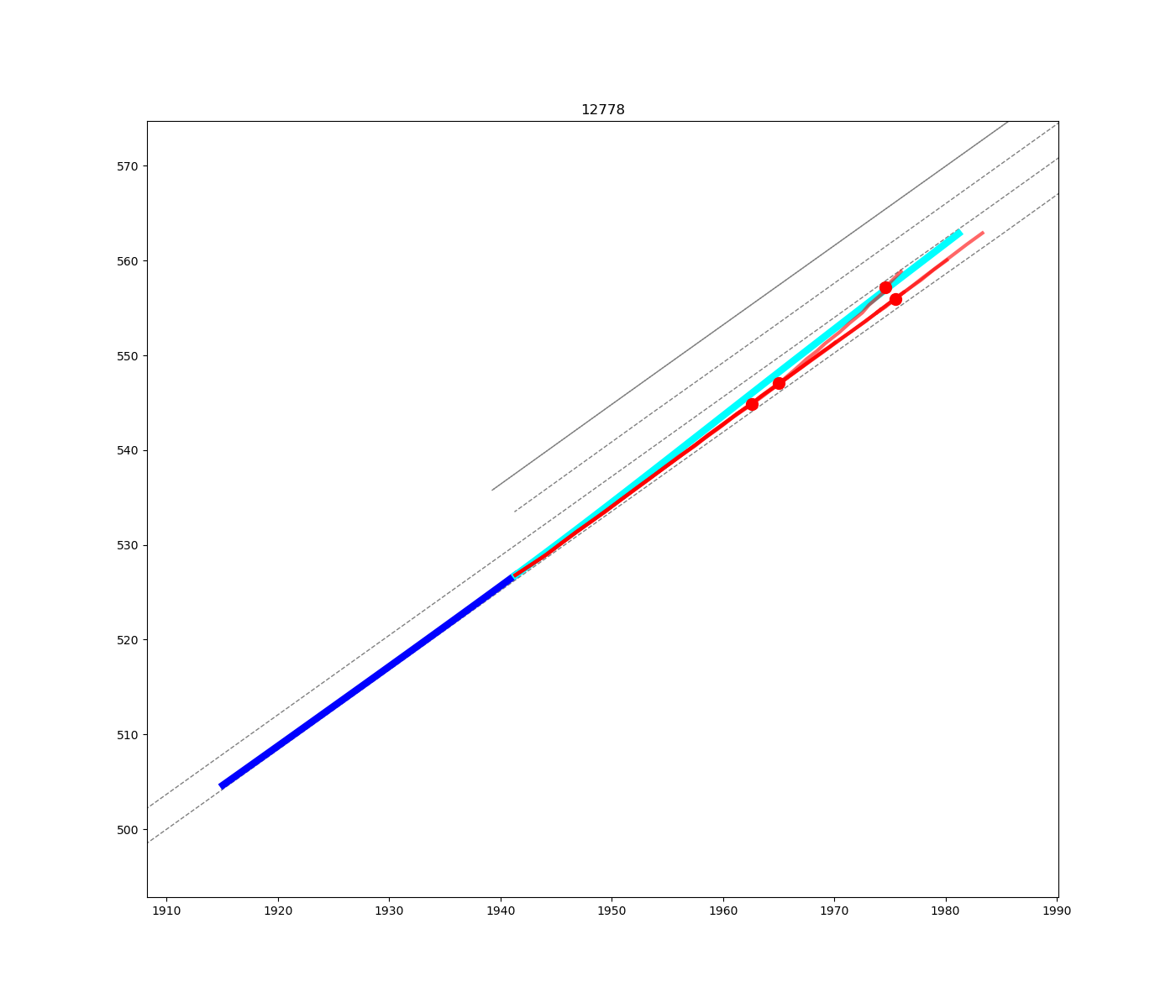}
        \includegraphics[trim=100 50 100 100,clip,scale=0.3]{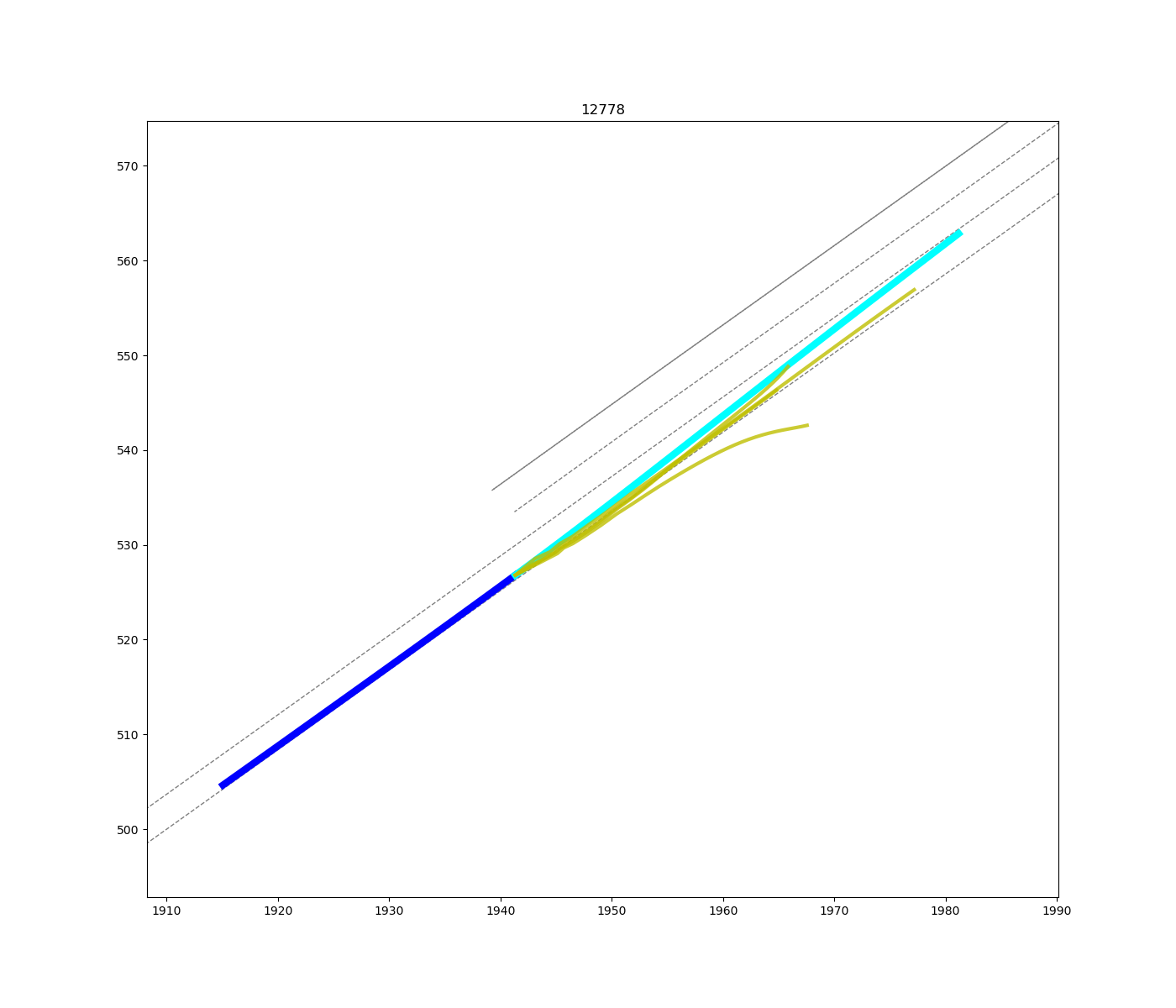}
\caption{Lane change and follow. Ground truth past and future trajectory are in blue and cyan. Top: Predictions from \modelname in red. Red dots depict the mode changes in predictions. Bottom: Predictions from ManeuverLSTM in olive.}
    \end{figure}
\newpage
\subsection{Lane follow and left turn}
\modelname (top) predicts an early left turn and a late left turn, after following the lane for a few seconds. ManeuverLSTM (bottom) predicts a left turn that does not account for mode change.

\begin{figure}[b!]
   \centering
        \includegraphics[trim=100 50 100 100,clip,scale=0.3
        ]{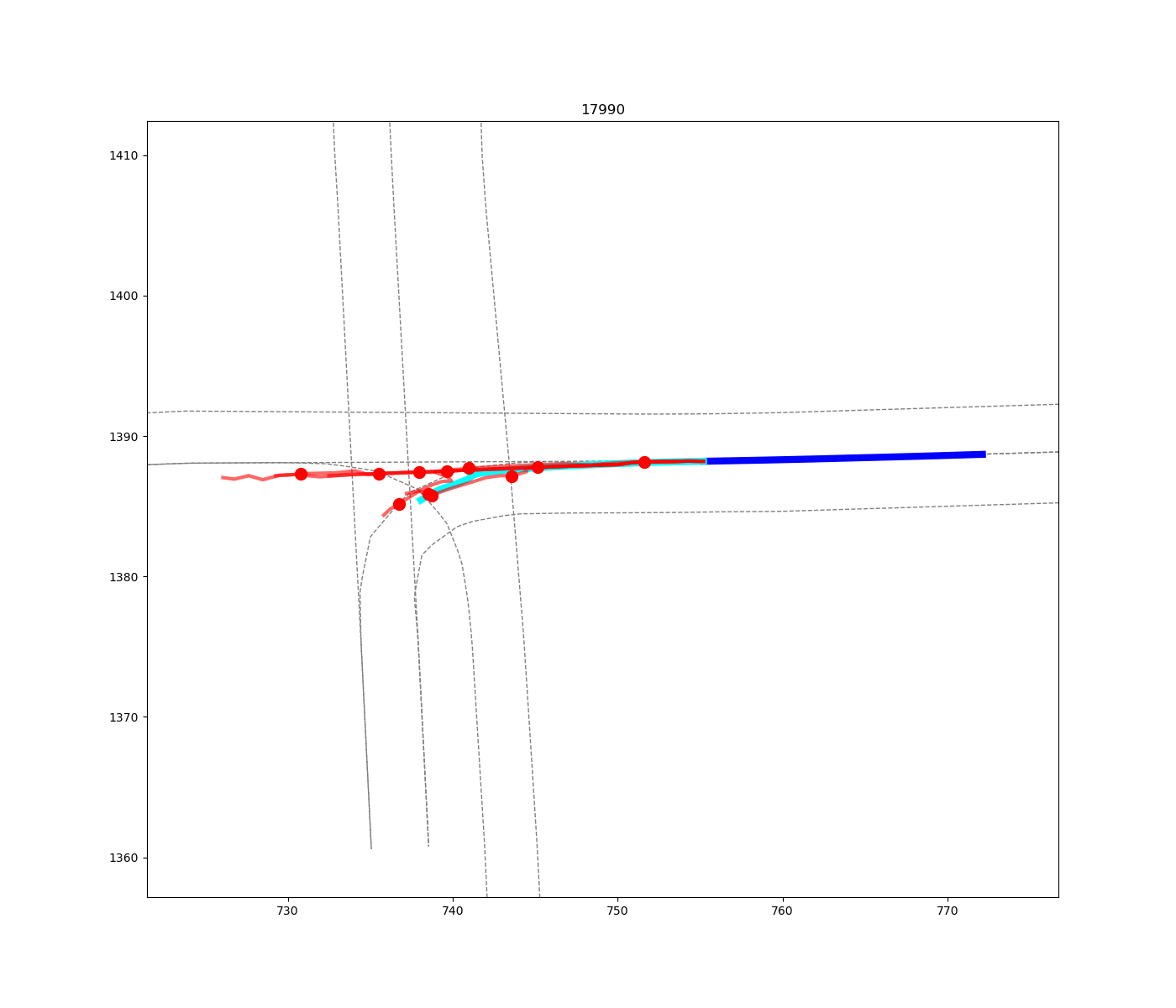}
        \includegraphics[trim=100 50 100 100,clip,scale=0.3]{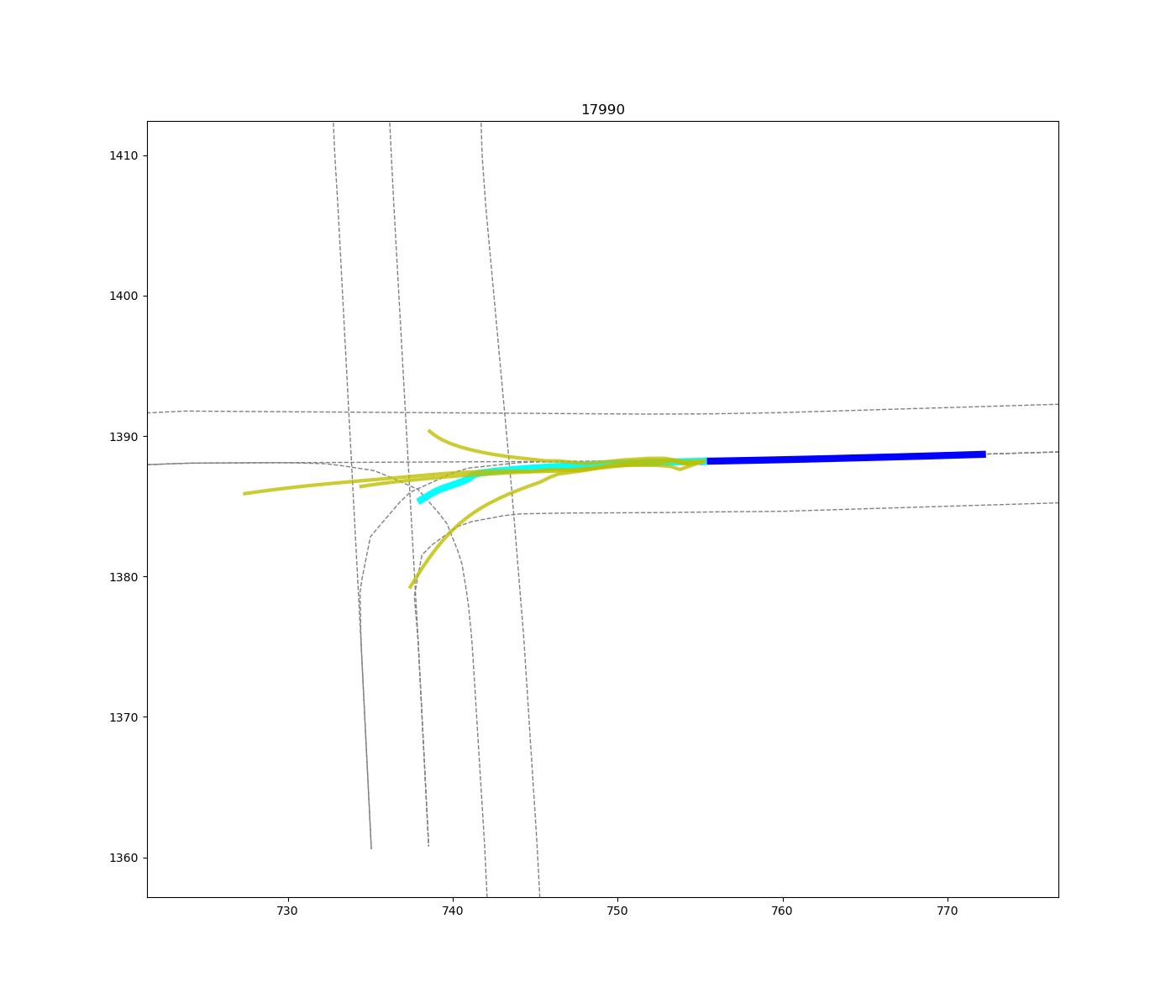}
\caption{Lane follow and left turn. Ground truth past and future trajectory are in blue and cyan. Top: Predictions from \modelname in red. Red dots depict the mode changes in predictions. Bottom: Predictions from ManeuverLSTM in olive.}
    \end{figure}
\newpage
\subsection{Lane follow and right turn}
\modelname (top) predicts the agent to follow the lane for a few seconds and then turn right. ManeuverLSTM (bottom) predicts a right turn that does not account for mode change.

\begin{figure}[b!]
   \centering
        \includegraphics[trim=100 50 100 100,clip,scale=0.3]{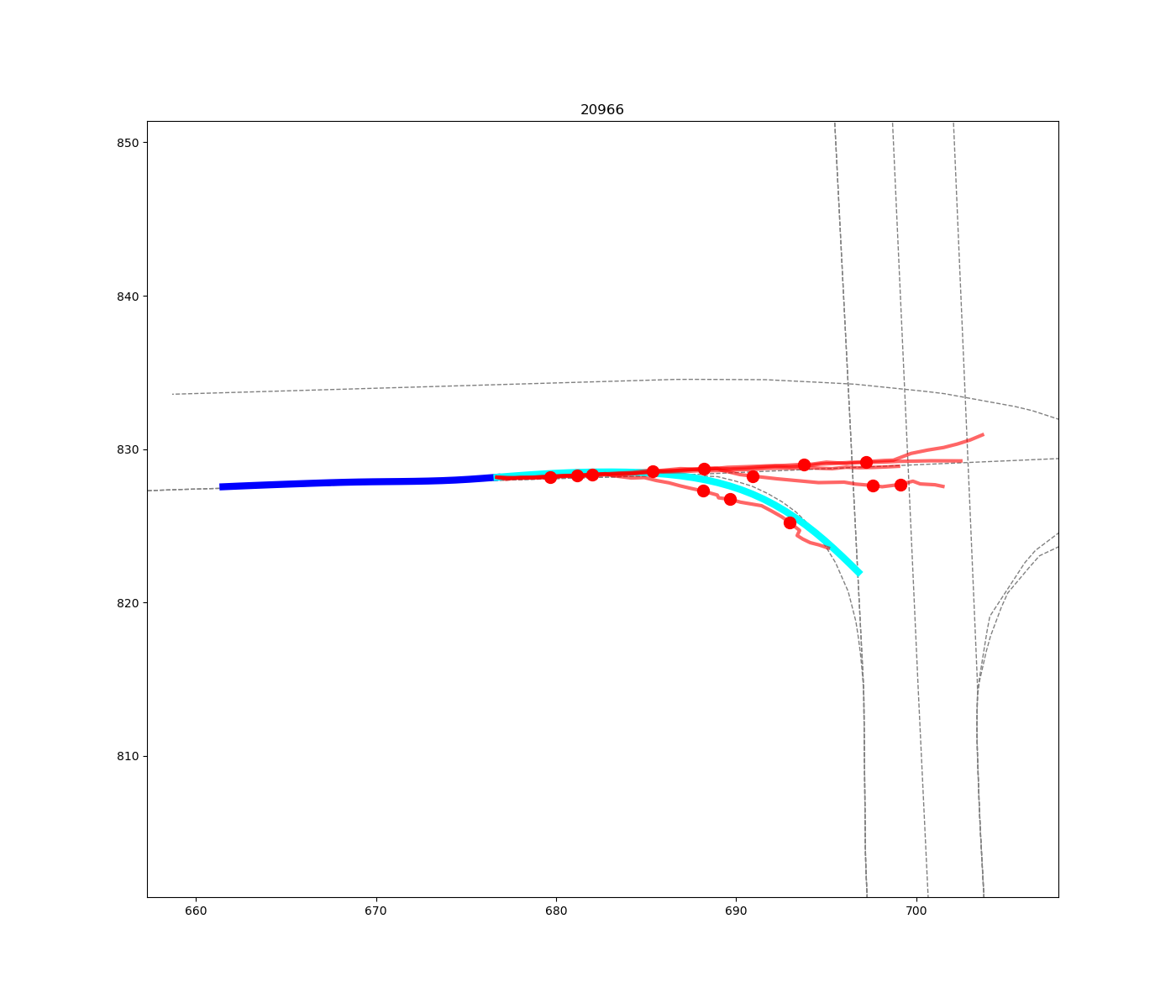}
        \includegraphics[trim=100 50 100 100,clip,scale=0.3]{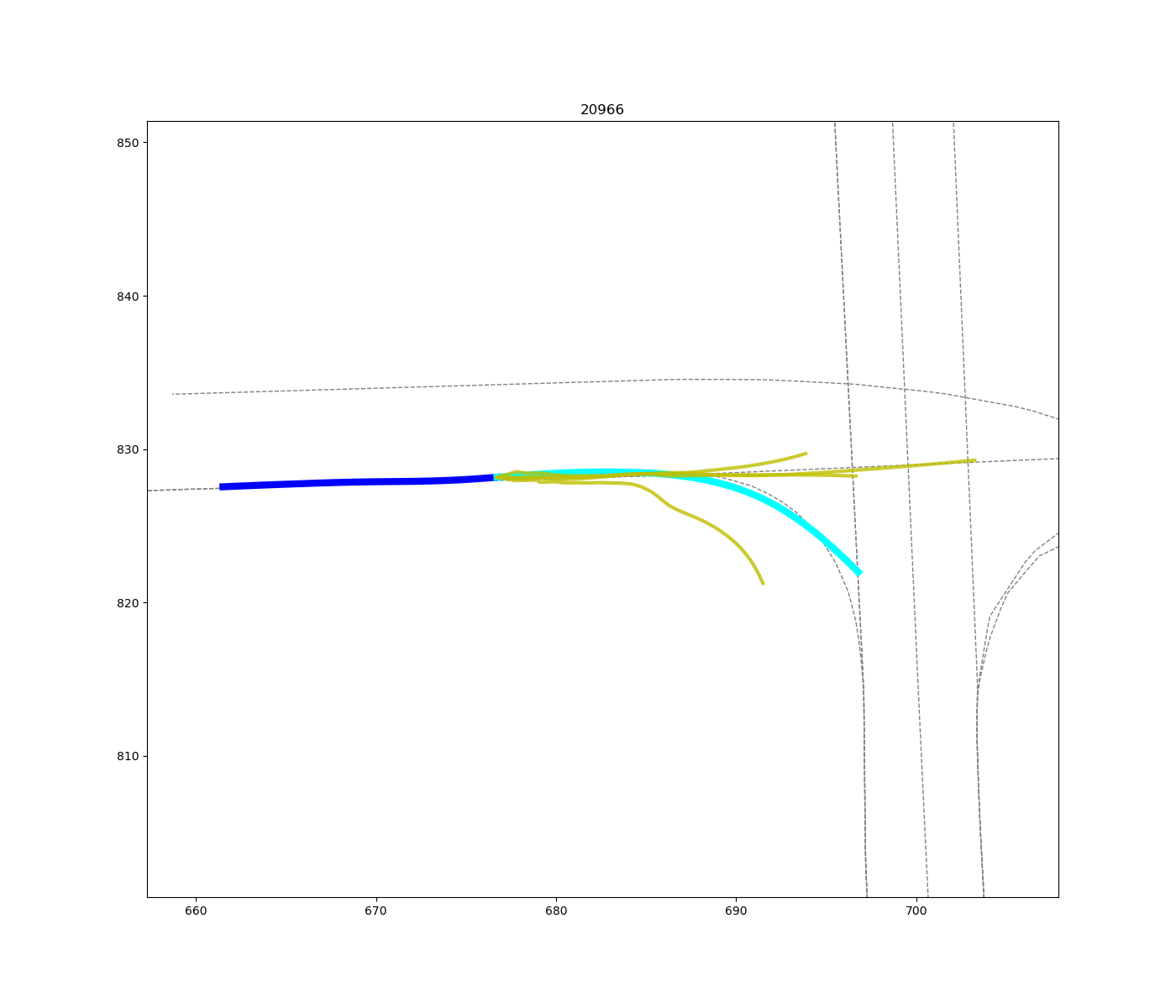}
\caption{Lane follow and right turn. Ground truth past and future trajectory are in blue and cyan. Top: Predictions from \modelname in red. Red dots depict the mode changes in predictions. Bottom: Predictions from ManeuverLSTM in olive.}
    \end{figure}
\newpage
\subsection{Lane follow and sharp turn}
\modelname (top) predicts a sharp right turn after following the lane for a few seconds. Although ManeuverLSTM achieves better final displacement error, it fails to account for the evolving intent and results in worse average displacement error. This issue is common in goal-conditioned models, which produce predictions with low final displacement errors by predicting goal targets explicitly, but suffers from high average displacement errors, as there are multiple ways to get to the goal.

\begin{figure}[b!]
   \centering
        \includegraphics[trim=100 50 100 100,clip,scale=0.3]{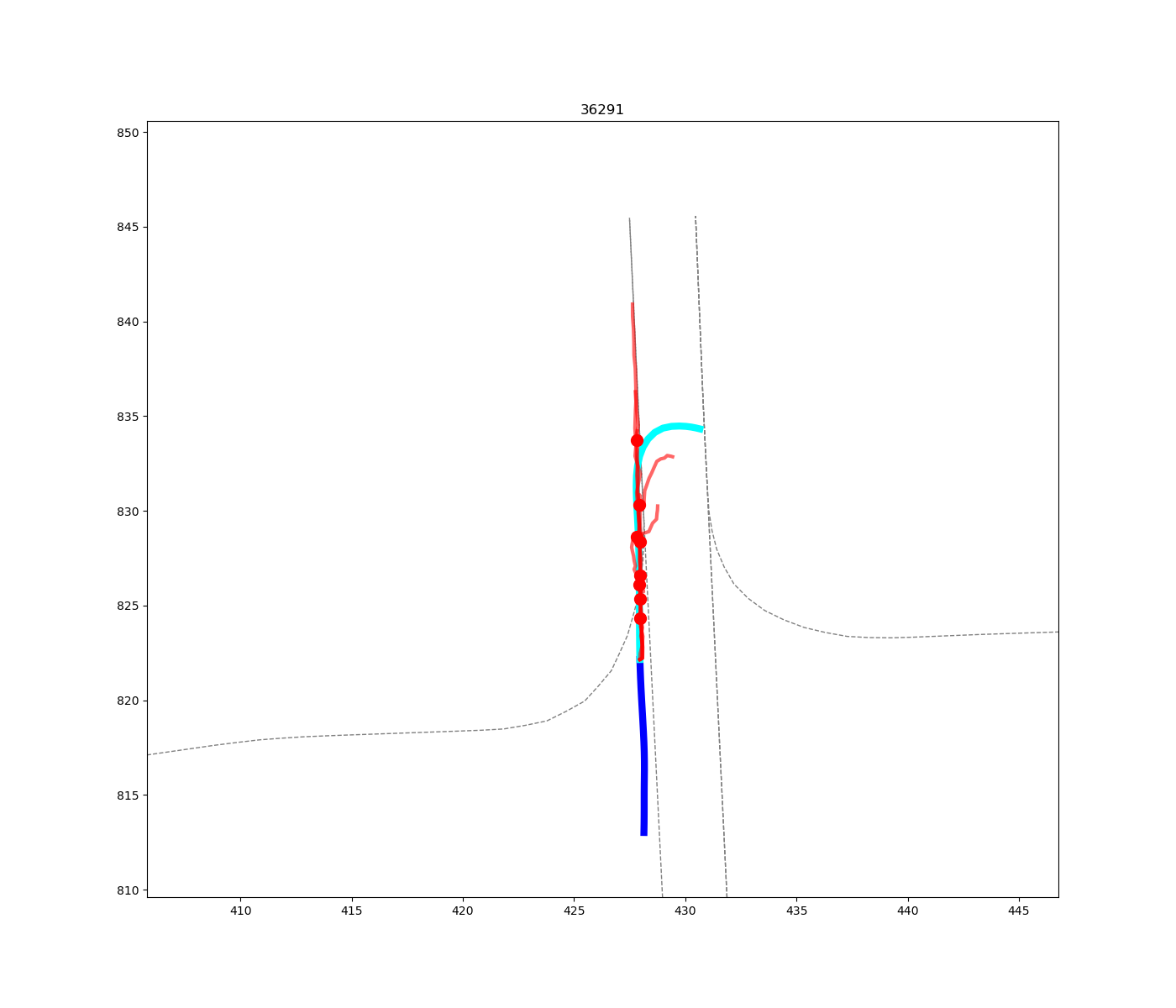}
        \includegraphics[trim=100 50 100 100,clip,scale=0.3]{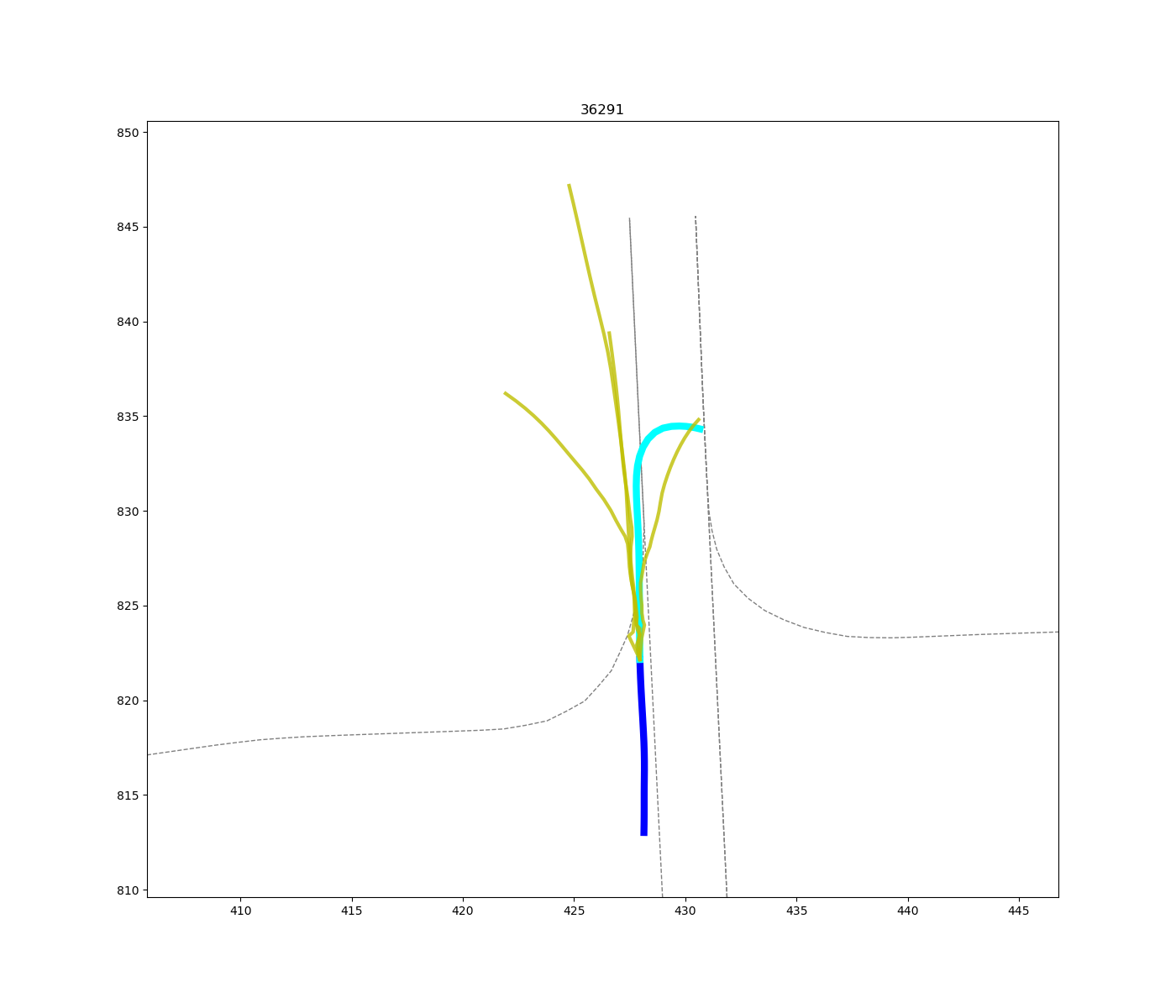}
\caption{Lane follow and sharp turn. Ground truth past and future trajectory are in blue and cyan. Top: Predictions from \modelname in red. Red dots depict the mode changes in predictions. Bottom: Predictions from ManeuverLSTM in olive.}
    \end{figure}
\newpage
\subsection{Fast froward and slow forward}
\modelname (top) predicts the agent to move forward fast and then slow down when approaching an intersection. 

\begin{figure}[b!]
   \centering
        \includegraphics[trim=100 50 100 100,clip,scale=0.3]{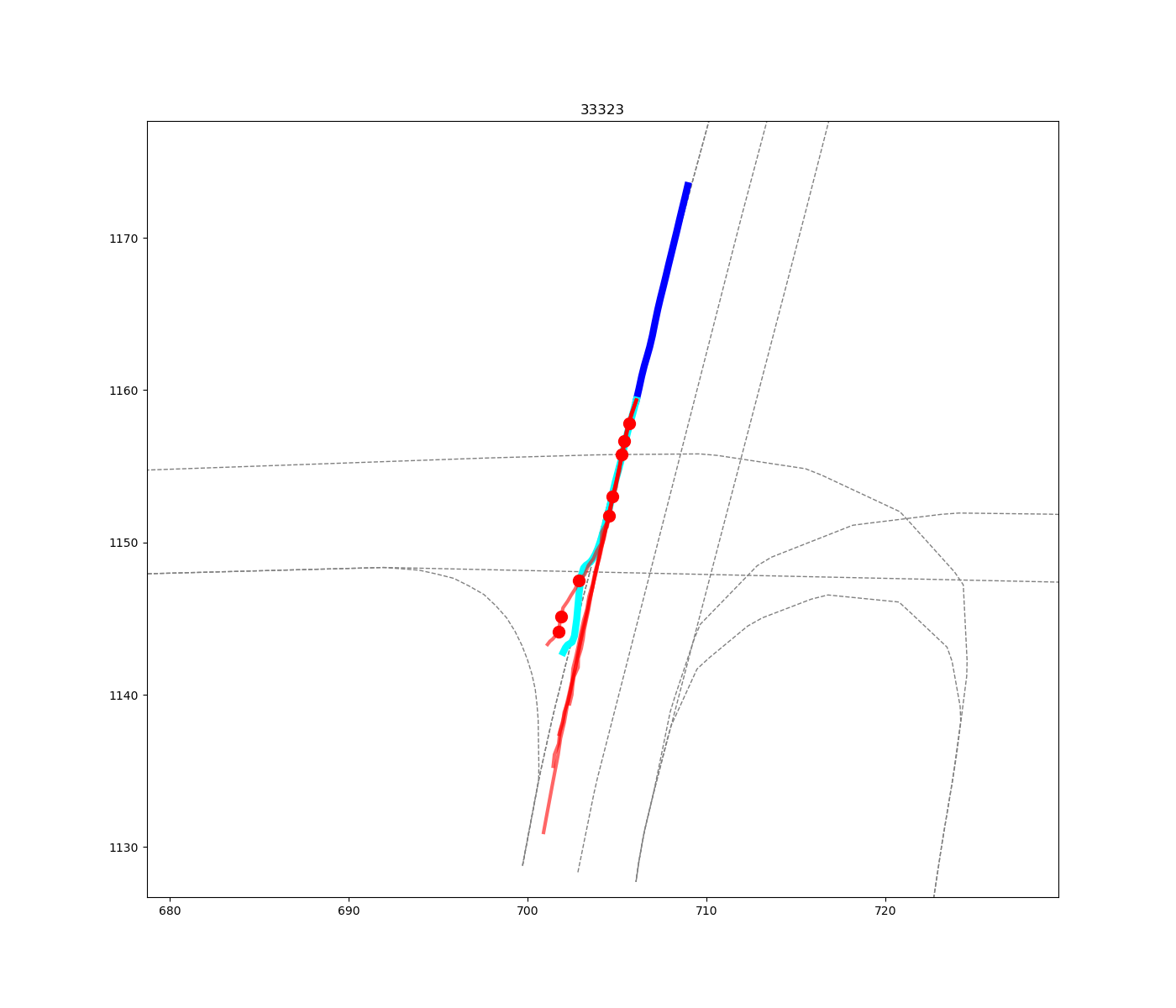}
        \includegraphics[trim=100 50 100 100,clip,scale=0.3]{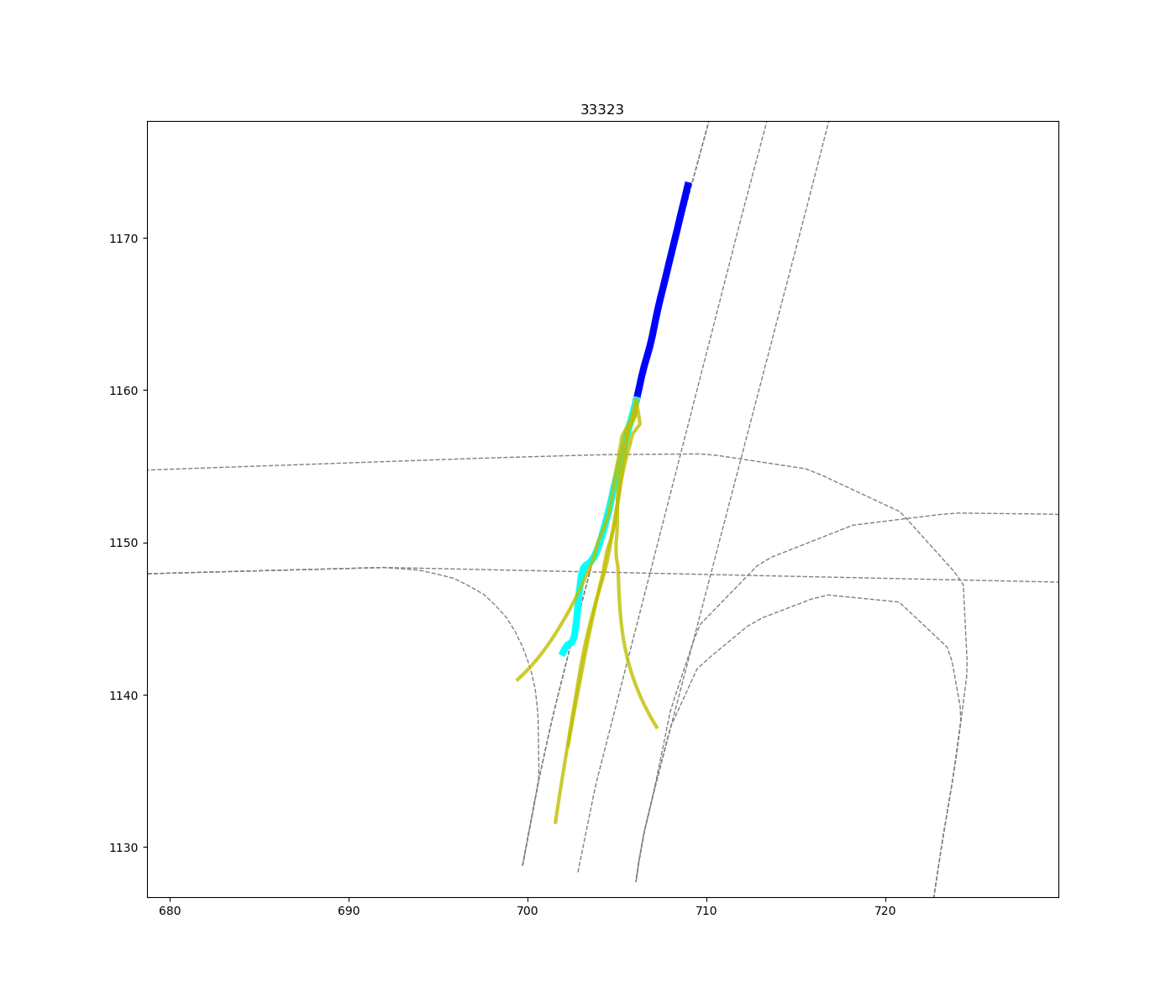}
\caption{Fast froward and slow forward. Ground truth past and future trajectory are in blue and cyan. Top: Predictions from \modelname in red. Red dots depict the mode changes in predictions. Bottom: Predictions from ManeuverLSTM in olive.}
    \end{figure}
\fi

\end{document}